\documentclass{article}

\usepackage{algorithm}
\usepackage{geometry}
\usepackage{authblk}
\usepackage{algorithmic}
\usepackage{microtype}
\usepackage{graphicx}
\usepackage{subcaption}
\usepackage{booktabs} 
\usepackage{hyperref}
\usepackage{amsmath}
\usepackage{amssymb}
\usepackage{mathtools}
\usepackage{amsthm}
\usepackage{tabularx}
\usepackage[capitalize,noabbrev]{cleveref}
\usepackage[textsize=tiny]{todonotes}
\usepackage{xltabular}
\usepackage{soul}
\usepackage{bm}
\usepackage{xspace}
\usepackage{multirow}
\usepackage{siunitx}
\usepackage[T1]{fontenc}
\usepackage{enumitem}
\usepackage{comment}
\usepackage{epstopdf}
\usepackage[nocompress]{cite}
\usepackage{caption} 
{}

{}
\newcommand{\final}[1]{#1}
\newcommand{\name}{{\sc Embroid}\xspace}
\newcommand{\ntasks}{{95}\xspace}

\newcommand{\shrink}{\tau}
\newcommand{\llm}{\lambda_{\sf LLM}}

\newcommand{\prompt}{\phi}

\newcommand{\data}{\mathcal{D}}
\newcommand{\x}{x}
\newcommand{\y}{y}

\newcommand{\nul}{n_u}
\newcommand{\prob}{\mathbb{P}}
\newcommand{\nemb}{N}
\newcommand{\EmbS}{\mathcal{E}}
\newcommand{\emb}{E}
\newcommand{\X}{\mathcal{X}}
\newcommand{\Y}{\mathcal{Y}}
\newcommand{\Z}{\mathcal{Z}}

\newcommand{\nlf}{m}

\newcommand{\lambdab}{\bm{\lambda}}

\newcommand{\nnx}{\text{NN}}
\newcommand{\svote}[1]{\lambda_{{\sf sm}, #1}}
\newcommand{\svotes}{\lambdab_{{\sf sm}}}

\newcommand{\sparam}{\alpha}

\newcommand{\E}[2]{\mathbb{E}_{#1}\left[#2\right]}
\newcommand{\Ehat}[1]{\hat{\mathbb{E}}\left[#1\right]}

\newcommand\independent{\protect\mathpalette{\protect\independenT}{\perp}}
\def\independenT#1#2{\mathrel{\rlap{$#1#2$}\mkern2mu{#1#2}}}

\theoremstyle{plain}
\newtheorem{theorem}{Theorem}[section]
\newtheorem{proposition}[theorem]{Proposition}

\theoremstyle{definition}

\theoremstyle{remark}

\title{Embroid: Unsupervised Prediction Smoothing Can Improve Few-Shot Classification}

\author[1]{Neel Guha*}
\author[1]{Mayee F. Chen*}
\author[1]{Kush Bhatia*}
\author[2]{Azalia Mirhoseini}
\author[3]{Frederic Sala}
\author[1]{Christopher Ré}

\affil[1]{Department of Computer Science, Stanford University}
\affil[2]{Anthropic}
\affil[3]{Department of Computer Science, University of Wisconsin-Madison}

\begin{document}

\maketitle
\def\thefootnote{*}\footnotetext{These authors contributed equally to this work.}\def\thefootnote{\arabic{footnote}}
\begin{abstract}
Recent work has shown that language models' (LMs) prompt-based learning capabilities make them well suited for automating data labeling in domains where manual annotation is expensive. The challenge is that while writing an initial prompt is cheap, improving a prompt is costly---practitioners often require significant labeled data in order to evaluate the impact of prompt modifications. Our work asks whether it is possible to improve prompt-based learning \textit{without} additional labeled data. We approach this problem by attempting to modify the predictions of a prompt, rather than the prompt itself. Our intuition is that accurate predictions should also be consistent: samples which are similar under some feature representation should receive the same prompt prediction. We propose \name, a method which computes multiple representations of a dataset under different embedding functions, and uses the consistency between the LM predictions for neighboring samples to identify mispredictions. \name then uses these neighborhoods to create additional predictions for each sample, and combines these predictions with a simple latent variable graphical model in order to generate a final corrected prediction. In addition to providing a theoretical analysis of \name, we conduct a rigorous empirical evaluation across six different LMs and up to 95 different tasks. We find that (1) \name substantially improves performance over original prompts (e.g., by an average of 7.3 points on GPT-JT), (2) also realizes improvements for more sophisticated prompting strategies (e.g., chain-of-thought), and (3) can be specialized to domains like law through the embedding functions.
\end{abstract}

\section{Introduction}\label{sec:introduction}

\begin{figure*}[th!]
    \centering
    \includegraphics[width=0.8\textwidth]{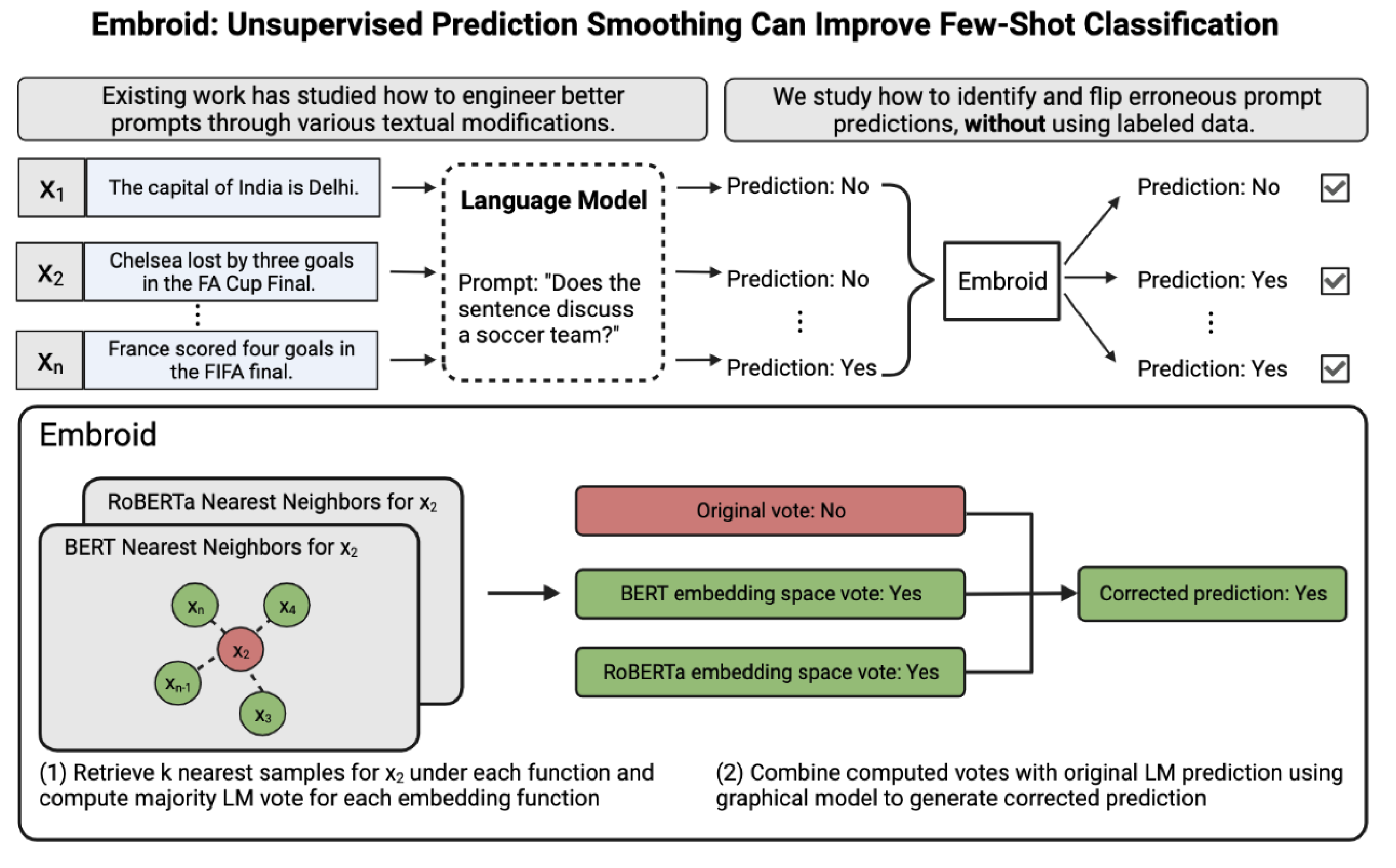}
    \caption{The \name method for prompt-patching. }
    \label{fig:banner}
\end{figure*}

Acquiring labeled data for domains like medicine and law is essential to training machine learning models or performing basic data analysis (e.g., ``how many contracts contain a choice-of-forum clause" or ``how many patient medical histories discuss an adverse reaction to a drug?'')~\cite{meerkat, guha2022legalbench}. However, building large labeled datasets is difficult, and efforts like \cite{cuad} show that  manual labeling with domain experts is cost-prohibitive. Recent works have begun exploring if language models (LMs) could learn annotation tasks \textit{in-context}~\cite{brown2020language} and replace manual labeling at scale~\cite{ding2022gpt, kuzman2023chatgpt, he2023annollm, meerkat}. The promise of this approach is that LMs' in-context capabilities enable them to learn tasks from descriptions of the the task (i.e., \textit{prompts}). However, the challenge is that producing high performance prompts is still expensive, as practitioners require labeled data in order to measure the impact of modifications to a prompt~\cite{perez2021true}. Existing work has thus focused on how domain experts can optimally construct prompts for a task (\textit{prompt-engineering}), while minimizing reliance on labeled data~\cite{liu2023pre, wei2022chain, su2022selective}. Yet, because language models are sensitive to even small changes in prompt language, these techniques are imperfect and still produce erroneous predictions~\cite{perez2021true, liang2022holistic, zhao2021calibrate, chen2022relation, arora2022ask}.

Our work approaches the challenge of improving prompt performance without labels from an orthogonal perspective: given the predictions of any prompted LM, can we identify and correct mis-predictions using \textit{unlabeled} data? We describe this as the problem of \textit{prompt-patching}. In the context of data annotation tasks, prompt-patching methods should meet three goals. First, they should be theoretically explainable, so that practitioners can understand when and how to apply them. 
Second, they should be fast, so that practitioners can efficiently integrate them into existing workflows. Finally, they should rarely be wrong, so that they don't worsen the performance of predictions.

Our work presents \name: a method for automatically identifying and correcting LM predictions with unlabeled data and no expert supervision. Recent work has shown that for many tasks, samples close-by in embedding spaces (produced by models like BERT) have the same label~\cite{chen2022shoring}. \name applies this intuition to the prompt-patching regime. Specifically, after LM predictions for all samples have been generated, \name retrieves the $k$ most similar samples for each input, under $N$ different embedding functions. For each embedding function, \name computes a scaled-modified majority vote over the LM's predictions for the $k$ retrieved samples. \name then combines these $N$ votes with the original LM prediction for the test sample using a simple latent variable graphical model that is learned with a fast method-of-moments estimator~\cite{fu2020fast}. The intuition behind \name is that good prompts are \textit{smooth} with respect to their predictions over a dataset---samples which are proximate under an embedding function should receive consistent predictions. Thus, modifying the predictions of a prompt to increase neighborhood agreement can improve the accuracy of those predictions. Lastly, because a single embedding space may imperfectly capture similarities between samples, retrieving neighbors from multiple embedding spaces improves robustness~\cite{jin2019probing, lin2019situating}.

Because \name relies on weak-supervision---the subject of recent rigorous study~\cite{chen2022shoring}---it is possible to theoretically analyze and explain \textit{why} and \textit{when} \name will improve performance. In particular, we find that  performance is a function of the quality of the embeddings and the performance of the initial prompt.  We also empirically study \name, conducting experiments over six LMs, on up to 95 tasks, with several different prompt strategies. We find that \name rarely worsens performance, and often improves F1 by a substantial margin. For instance, \name improves GPT-3.5 by an average of 4.9 points F1 per task, and GPT-JT by an average of 7.3 points per task. The magnitude of \name's gains are such that it enables a 1.3B parameter model to outperform an instruction-tuned 6.7B parameter model. \name is also complementary to advanced prompt engineering strategies, and achieves performance improvements when applied to prompts designed using chain-of-thought~\cite{wei2022chain}, AMA~\cite{arora2022ask}, and selective annotation~\cite{su2022selective}. Finally, \name can be extended to specialized domains like law, through the use of already-available domain specific embeddings.

Succinctly, our contributions in this paper are: (1) \name, a simple prompt-patching framework for improving LM predictions over text classification tasks; (2) a theoretical analysis of \name which explains performance improvements in terms of embedding smoothness and base accuracy; and (3) an empirical evaluation of \name covering up to 95 tasks and six different LMs.
\section{Related work}\label{sec:rel}

\paragraph{Improving LM performance} Improving the in-context generalization abilities of LMs has been intensely studied. The first family of approaches focuses on adapting LMs in order to make them more amenable to prompting. This includes task-specific finetuning~\cite{nakamura2019revisiting, hsieh2023distilling, huang2022large}, training on instruction data~\cite{alpaca, chung2022scaling}, RLHF~\cite{ouyang2022training}, and weight-surgery methods which attempt to ``correct'' incorrect information stored in model weights~\cite{meng2022locating, meng2022mass, hase2021language, dai2021knowledge}. A second family of approaches explores strategies for optimizing prompts to models, either through the specific textual features of the prompt~\cite{wei2022chain, jung2022maieutic, mishra2021reframing}, the use of task decompositions or LM recursion~\cite{arora2022ask},  implicit prompt representations~\cite{li2021prefixtuning, lester2021power}, or external databases~\cite{mitchell2022memory}.  Prompt-patching, in contrast, focuses on identifying mistakes in the predictions generated from a particular prompt. The most related approaches are aggregation methods, in which the outputs of multiple prompts are combined with an ensembling method~\cite{liu2023pre, arora2022ask}. We find that \name outperforms many such baselines, and can be applied to enhance their outputs. 

\paragraph{Weak supervision} \name leverages statistical techniques developed in the weak supervision literature. The objective in weak supervision is to generate probabilistic labels for unlabeled data by combining the predictions of multiple noisy heuristics~\cite{ratner2017snorkel, ratner19, fu2020fast, shin22, Wu22, vishwakarma2022lifting}. \name's novelty is that it uses embeddings to construct additional synthetic predictions, which are combined with the original predictions. In contrast, recent weak supervision approaches which incorporate embeddings use them to produce more fine-grained accuracy parameters~\cite{chen2022shoring}, detect and discard training points~\cite{lang2022training}, and as the basis for label propagation with final weak supervision predictions~\cite{pukdee2022label}.

\section{Problem setup and background}\label{sec:prob}

\paragraph{Problem setup} Our problem setup comprises three elements: an unlabeled dataset, predictions from a LM for each sample in this dataset, and embedding representations of our dataset. Our goal is to improve the accuracy of LM predictions, by using the embedding representations to identify and correct predictions likely to be incorrect. Because recent work has explored how predictions from multiple prompts can be combined for a task~\cite{arora2022ask}, we present a generalized version of \name in which we have access to multiple LM predictions. In our empirical evaluation however, we show that \name performs well regardless of the number of predictions per sample available.

More formally, we focus on a binary classification task where $\small{x \in \X}$ denotes a sentence or paragraph and $\small{y \in \Y = \{-1, 1\}}$ is the binary label. We assume we are given an unlabeled dataset $\small{\data = \{\x_i\}_{i = 1}^{\nul}}$ of $\small{\nul}$ points. Each point $x$ is sampled i.i.d. from a distribution $\small{\prob_x}$, and there exists a true underlying distribution $\prob$ on the joint $\small{(\x, \y)}$. Following the true few-shot regime~\cite{perez2021true}, we assume the only labels available are those used in the prompt. We denote a language model (e.g., GPT-3) as $\small{\llm}$, and a task-specific prompt as $\small{\prompt}$, which prepends task instructions to input $x$ (e.g., ``Does the clause contain an audit provision? Yes or No."). The prediction this prompt induces for $\small{\llm}$ over $\x$ is $\small{\llm(\prompt(\x)) \in \Y}$.\footnote{We assume a task-specific mapping function which allows a practitioner to associate a text generation from an LM to a particular class prediction in $\Y$.} Varying $\small{\prompt}$ by changing the task description, in-context demonstrations, or punctuation will alter the prediction generated for $x$. For a set of $m$ prompts $\small{[\prompt_1, \dots, \prompt_m]}$, we denote their respective  predictions on $x$ as a vector of \textit{weak sources} $\small{\lambdab(x) = [\llm(\prompt_1(x)), \dots, \llm(\prompt_m(x))]}$. For convenience, we denote $\small{\llm(\prompt_i(x))}$ as $\small{\lambda_i(x)}$ or $\small{\lambda_i}$ when the $x$ is obvious, and similarly use $\small{\lambdab}$ instead of $\small{\lambdab(x)}$. We distinguish between two regimes: in the \emph{single-prompt} regime with $m = 1$, we have access to a LM prediction for each point $\x$, while in the \emph{multi-prompt} regime with $m > 1$, we have access to multiple predictions.

We assume access to $\small{\nemb}$ embedding models $\small{\EmbS = [\emb_1, \ldots, \emb_\nemb]}$, each represented as a fixed mapping $\emb_i : \X \mapsto \Z_i$ from an input $\x$ to an embedding vector $z$. These auxiliary embedding models provide representations of $x$ which encode different types of similarity information. Through model repositories like HuggingFace~\cite{wolf2020transformers}, it is possible to download a number of models which generate representations for text sentences (e.g., BERT or RoBERTa~\cite{devlin2018bert, liu2019roberta}). These models have the property that semantically similar sentences are close-by in embedding space~\cite{chen2022shoring, jin2019probing, lin2019situating}.

\paragraph{Weak supervision background} Weak supervision uses a graphical model to combine votes from multiple noisy sources into a single prediction, by estimating the accuracy of each source. It models $\small{\Pr(y, \lambdab(x))}$ as a latent variable graphical model and uses $\small{\hat{y} = \text{argmax}_y \hat{\Pr}(y | \lambdab(x))}$ to produce label estimates, where $\hat{\Pr}$ represents the learned  model. The graphical model is based on a graph $\small{G = (V, E)}$, where $\small{V = y \cup \lambdab}$ and $E$ consists of edges from $y$ to each $\lambda_j$. We assume no dependencies between sources, although simple extensions can incorporate them~\cite{varma2019learning}. The formal graphical model is:
\begin{align}\label{eq:base}\small{
    \Pr(y, \lambdab(x)) = \frac{1}{Z} \exp (\underbrace{\theta_y y}_{(I)} + \underbrace{\theta^\top \lambdab(x) y}_{(II)})
    }
\end{align}

where $Z$ is the partition function used for normalization, $(I)$ represents a label balance term with parameter $\theta_y$ controlling the prior of $\small{\Pr(y = 1)}$, and $(II)$ represents the source accuracy term where each $\theta_i$ is an \textit{accuracy parameter} for the $i$th source. Note that from this model, sources are conditionally independent: $\small{\lambda_i \independent \lambda_j | y}$ for any $i, j \in [m]$. Our use of this model has two steps. First, we must learn the accuracy parameters of $\Pr(y, \lambdab(x))$ without access to $y$.  We use the triplet method introduced in~\cite{fu2020fast}, which is an efficient method-of-moments estimator for the parameters. Then, at inference we compute $\small{\hat{\Pr}(y | \lambdab(x))}$. Appendix~\ref{app:wsback} contains more details.
\section{\name}
\begin{algorithm}[tb]
   \caption{\name: Correcting LLMs with embeddings}
   \label{alg:embroid}
\begin{algorithmic}
   \STATE {\bfseries Input:} Unlabeled data $\data$, LLM predictions $\lambdab(x)$ for each $x \in \data$, 
   embedding models $\EmbS = \{\emb_1, \ldots, \emb_\nemb\}$, shrinkage parameter $\shrink$, nearest neighbors parameter $k$

   \FORALL{unlabelled $x \in \data$}
   \FORALL{embedding models $\emb_j \in \EmbS$}
   \STATE Compute k-nearest neighbors $\nnx_{j, k}(x)$
   \STATE Compute smoothed neighborhood prediction $\svote{j}(x)$ using $\lambdab$, $\nnx_{j, k}(x)$, and $\tau$ using eq.~\eqref{eq:svote}
   \ENDFOR
   \ENDFOR
   \STATE Solve graphical model $\Pr(y, \lambdab(x), \svotes(x))$ in eq.~\eqref{eq:main} with triplet method over $\data$ (Algorithm~\ref{alg:triplet}).
   \FORALL{unlabeled $x \in \data$}
   \STATE Sample $\hat{\y}_x \sim \hat{\Pr}(y|\lambdab(x),\svotes(x))$
   \ENDFOR
   \STATE {\bfseries Output:} Label set $\bm{\hat{Y}} = \{\hat{\y}_x \, | \, x \in \data\}$

\end{algorithmic}
\end{algorithm}

First, \name uses the embedding models $\small{\EmbS}$ to compute additional votes for each $\x$. Let $\small{\nnx_{j,k}(x) \subset \data}$ be the $k$-nearest neighbors of sample $x$ under the embedding function $\small{\emb_j}$. We define the smoothed neighborhood prediction vector $\small{\svote{j}(x) \in \{-1, 0, 1\}^m}$ as follows, with $\small{\svote{j}[i](x)}$ being the $i$th element:

{\small
\begin{align}
\label{eq:svote}
\begin{gathered}
\tilde{\lambda}_j[i](x) = \frac{1}{k} \sum_{\tilde{x} \in \nnx_{j,k}(x)} \lambda_i(\tilde{x})  \\
\svote{j}[i](x) = \begin{cases}
1 & \tilde{\lambda}_j[i](x) > \shrink^{+}_i\\
-1 & \tilde{\lambda}_j[i](x) < \shrink^{-}_i\\
0 & \text{o.w.}
\end{cases}\;,
\end{gathered}
\end{align}
}
where $\small{\shrink^{+}_i \in [-1, 1]}$ and $\small{\shrink^{-}_i \in [-1, 1]}$ act as shrinkage parameters for $\lambda_i$ which control the level of agreement amongst the neighbors of $x$ necessary to generate a particular vote. The scalar $\small{\svote{j}[i](x)}$ is the average vote of $\lambda_i$ amongst the neighbors of $x$ in $\emb_j$. When $\small{\svote{j}[i](x)}$ is sufficiently positive, i.e., $\small{\svote{j}[i](x)} > \tau^{+}_i$, \name sets $\small{\svote{j}[i](x)}$ to be a positive vote. When $\small{\svote{j}[i](x)}$ is sufficiently negative, i.e., $\small{\svote{j}[i](x)} < \tau^{-}_i$, \name sets $\small{\svote{j}[i](x)}$ to be a negative vote. Otherwise, $\small{\svote{j}[i](x)}$ is set to be an abstain. The intuition is that $\small{\svote{j}[i](x)}$ will be an accurate vote over $x$ whenever two conditions are met: (1) the LM is generally accurate, i.e., $\lambda_j$ is usually correct, and (2) $\emb_j$ is \textit{smooth}, i.e., nearest-neighbors share the same task label.

Next, we augment our base model in equation~\eqref{eq:base} to incorporate these auxiliary neighborhood predictions $\small{\svotes = [\svote{1}, \dots, \svote{N}] \in \{-1, 0, 1\}^{\nemb \nlf}}$ computed using the embeddings:
{\small
\begin{align}\label{eq:main}
&\Pr(y, \lambdab, \svotes) = \frac{1}{Z}\exp\Bigg(\theta_y\y + \theta^\top \lambdab \y + \sum_{j=1}^\nemb \sparam_j^\top \svote{j}\y\Bigg),
\end{align}
}
where the vector $\alpha_j \in \mathbb{R}^{\nlf}$ represents the quality parameters for the $j^{th}$ embedding model when used with the $\nlf$ different prompts. To solve this model and produce label estimates, we note that it has the same format as~\eqref{eq:base} if we concatenate $\lambdab$ and $\svotes$ into one set of weak sources. Therefore, we can use the triplet method  from~\cite{fu2020fast} to learn parameters and output estimates $\small{\hat{\Pr}(y | \lambdab(x), \svotes(x))}$ for each $x \in \data$ at inference time (see Appendix~\ref{app:wsback} for details).

Parameters $\theta$ and ${\alpha_j}$ in~\eqref{eq:main} allow us to trade-off two different sources of information---one presented by directly prompting an LM to obtain a label and the other by incorporating similarity information from the embedding models---and to further account for varying error modes among the embedding models. Our use of the neighborhood predictions in~\eqref{eq:main} yields a more expressive model than the standard weak supervision framework solely on LLM predictions in~\eqref{eq:base}, which we can recover when $k = 0$, and can thus help make corrections to the LLM predictions. In practice, we find that setting $\tau^+_i = \tau^-_i = \mathbb{E}[\lambda_i]$ (i.e., the average source vote) yields good performance (Appendix \ref{sec:appendix:hyperparameters}).

\section{Theoretical analysis} \label{sec:theory}

We  analyze \name, discussing the advantages of using $\svotes$ in addition to $\lambdab$, and show that embedding smoothness and base prediction accuracy play a critical role in information gain. Appendix \ref{appendix:synthetics} provides synthetics demonstrating these tradeoffs and comparing to weak-supervision baselines.

First, we provide a result on the generalization error of our model $\small{\hat{\Pr}(y | \lambdab, \svotes)}$. Define the generalization error as the expected cross-entropy loss, $\small{L(\lambdab, \svotes, \data) = \mathbb{E}_{y, \lambdab(x), \svotes(x), \data}[-\log \hat{\Pr}(y | \lambdab(x), \svotes(x))]}$.
We use $\small{[\lambda_1, \dots, \lambda_{(\nemb + 1)\nlf}]}$ to represent  $\small{[\lambdab, \svotes]}$ and
denote by $\small{a_{\max} = \max_{i} \E{}{\lambda_i(x) y}}$ the largest accuracy (scaled to $[-1, 1]$) of any source, and by $\small{b_{\min} = \min_{i, j} \{\E{}{\lambda_i \lambda_j}, \Ehat{\lambda_i \lambda_j}\}}$ the minimum expected pairwise product between any two sources. Assume that all sources are better than random, e.g., $\Pr(\lambda_i = y) > 0.5$.  These terms and assumptions are from using the triplet method. 

\begin{proposition} \label{prop:gen_err}
Suppose that the data $x, y, \lambdab, \svotes$ follows the model in~\eqref{eq:main}. The generalization error of $\hat{\Pr}(y | \lambdab, \svotes)$ can be decomposed into 
\begin{align}
L(\lambdab, \svotes, \data) \le  \underbrace{H(y | \lambdab, \svotes)}_{\text{Irreducible Error}} + \underbrace{\frac{C (\nemb + 1)\nlf}{\nul}}_{\text{Variance}} + o(1/\nul), \nonumber 
\end{align}
where $C = \frac{3(1 - b_{\min}^2)}{8b_{\min}^2 (1 - a_{\max}^2)} \big(\frac{1}{b_{\min}^4} + \frac{2}{b_{\min}^2}\big)$.
\end{proposition}

In the bound above, the variance term  comes from estimation error when learning the parameters via the triplet method. The irreducible error depends on quality of $\lambdab$ and $\svotes$. If knowledge of the LLM prediction and neighborhood prediction significantly reduces uncertainty in $y$, the conditional entropy term $H(y | \lambdab(x), \svotes(x))$ is low. 

\paragraph{Information gain from using both $\lambdab, \svotes$}
We compare upper bounds on generalization error when both $\lambdab, \svotes$ are modeled versus when only $\lambdab$ is modeled, as in~\eqref{eq:base} corresponding to classical weak supervision. Based on the bound in Proposition~\ref{prop:gen_err}, modeling both $\lambdab$ and $\svotes$ increases the variance term by a constant multiplicative factor. 

Here, we examine how the irreducible error is affected, that is, the difference $H(y | \lambdab) - H(y | \lambdab, \svotes)$. Since this quantity is always nonnegative, we focus on bounding the \textit{pointwise} difference in conditional entropy---which we call the information gain---for a given $x_0$ on which the LLM is incorrect. For simplicity, suppose we have one embedding $\emb$. An embedding $\emb$ is $M$\textit{-smooth} with respect to the label if 
\begin{equation}\label{eq:smoothness}
\Pr(\tilde{y} = c | y = c, \|\emb(x) - \emb(\tilde{x}) \| \le \varepsilon) \ge M_{\emb}(\varepsilon),
\end{equation}
where $c \in \Y$, $\varepsilon > 0$ and $M_{\emb}(\cdot) \in [0, 1]$ is decreasing in its input. Define $\beta_i = \Pr(\lambda_i = y)$ as the accuracy of $\lambda_i$ and $p_{\lambdab} = \Pr(y = 1 | \lambdab(x_0))$ as the prediction on $x_0$ given only access to $\lambdab$. Let $\varepsilon_k = \max_{\tilde{x} \in \nnx_{k}(x)} \| \emb(x) - \emb(\tilde{x}) \|$ be the maximum distance between $x_0$ and its $k$ neighbors. Without loss of generality, assume the label on $x_0$ is $y = 1$.

\begin{theorem} \label{thm:cond_entr_smoothness}
Assume that $\emb$ is $M$-smooth. The pointwise information gain on $x_0$  is 
\begin{align}
    &H(y | \lambdab(x_0)) - H(y | \lambdab(x_0), \svotes(x_0)) \ge \nonumber \\
    &2(1 - p_{\lambdab}) \bigg[ \prod_{i = 1}^m \big[1 - \exp[-2k(\beta_{\nnx_k, i} - 0.5)^2]\big] - 0.5 \bigg] \nonumber
\end{align}

where $\beta_{\nnx_k, i} = \Pr_{\tilde{x} \sim \nnx_{k}}(\lambda_i(\tilde{x}) = y) \ge \beta_i M_E(\varepsilon_k)$ is the neighborhood accuracy.
\end{theorem}

A few observations on the bound are in order.
\begin{itemize}
    \item \textbf{Improvement over WS:} If the neighborhood accuracy is bounded sufficiently far from $\frac{1}{2}$ and $k$ is large, using \name has better irreducible error than just using $\lambdab$. 
    For example, setting $m = 1$, $k = 10$, $\beta_{\nnx_k, i} = 0.7$, and $p_{\lambdab} = 0.25$ gives us an improvement of $0.076$ nats.
    \item \textbf{Smoothness:} If $E$ is highly smooth, then $M_E(\varepsilon_k)$ will be large and irreducible error will be small.
    \item \textbf{Base prediction accuracy:} If the original sources $\lambdab$ have high accuracy ($\beta_i$), irreducible error will be small.
\end{itemize}

Additionally, we observe that if $p_{\lambdab}$ is a high-quality prediction close to the true label $1$, the information gain is small. Choice of the $k$ parameter presents a performance trade-off: increasing $k$ will increase $\varepsilon_k$ and incorporate farther-away, less reliable predictions, but it will also reduce the noise of the majority vote. We also comment on the information gain when using both $\lambdab$ and $\svotes$ over just $\svotes$ in Appendix~\ref{appendix:proofs}.
\section{Results}\label{sec:results}
Our empirical evaluation focuses on three questions: (1) How robust is \name's performance across LMs? (2) How does \name, as a prompt-patching method, compare to high performance prompt-engineering methods? (3) How sensitive is \name to the embeddings and dataset size?

\paragraph{Tasks} We study tasks where sentence embeddings can capture information relevant to the task, leading us to focus on sentence classification datasets. We consider a collection of 95 class-balanced sentence classification tasks, derived from binarizing existing multi-class legal, scientific, and general domain classification benchmarks like CUAD, AGNews, DBpedia-14, FewRel, and several others~\cite{cuad, chemprot, fewrel, wrench, zhang2015character}.\footnote{We hope to explore multi-class extensions and more complex reasoning tasks in future work.} Example tasks include, ``Classify if the following texts discuss a recording label'' or ``Classify if the following contractual clauses contain an audit rights provision.'' 

\paragraph{Choice of embedding models} Following prior work illustrating the benefits of domain specific representation~\cite{zheng2021does, gururangan2020don}, \name uses different embeddings for each task domain. For law tasks, we rely on two BERT-variants trained on different legal corpora~\cite{zheng2021does, henderson2022pile}. For science tasks, we rely on three BERT-variants trained on science, biology, and medical texts~\cite{pubmedbert, lee2020biobert, beltagy2019scibert}. For general domain tasks, we rely on BERT, Roberta, and SentenceBert embeddings~\cite{liu2019roberta, devlin2018bert, reimers2019sentence}. 

\paragraph{Prompts} Prompts are constructed using fixed instructions, and by manually selecting three random samples (from each class) as in-context demonstrations (Appendix \ref{appendix:prompts}). We follow the true few-shot regime~\cite{perez2021true}, in that we assume the only labeled data available to the data scientist are the labels used for in-context demonstrations. Prior work has found this regime to most realistically represent real-world workflows. 

\paragraph{Models} We evaluate on two API-access models: GPT-3.5 (\texttt{text-davinci-003}) and J1-Jumbo~\cite{lieber2021jurassic}. Because API models raise significant privacy and compliance concerns for data scientists working with sensitive data~\cite{bloomberg_employers}, we also evaluate on open-source models. We select models in the 6-7B parameter range, as these are the largest models which fit on commonly available 40GB A100 machines. Specifically, we evaluate Bloom~\cite{scao2022bloom} and OPT~\cite{zhang2022opt}. Given the increasing popularity of instruction-tuning, we also evaluate on GPT-JT~\cite{together-gptjt}, an 6.7B parameter instruction tuned version of GPT-J. Because of cost-constraints, we evaluate API-access models on a representative selection of 12 tasks, while evaluating all other models on the full suite of 95 tasks. Appendix~\ref{appendix:datasets} provides details.

\subsection{By how much does prompt-patching improve performance?}
\paragraph{Performance across LM families} We examine if \name achieves improvements for different \textit{types} of LMs. For each LM, we select three different combinations of in-context demonstrations (i.e., three prompts), generate predictions for each prompt, and apply \name to independently each prompt's predictions. This produces $3 \times 95 = 285$ trials for open-source models, and $3 \times 12 = 36$ trials for API-models. We report \textit{win-rate}, i.e., the proportion of trials for which \name outperforms the original predictions, and \textit{improvement}, i.e., the average difference in F1 points (across all trials) between \name and the original predictions. 

As Table \ref{tab:embroid_improvement} illustrates, \name improves performance for a substantial proportion of prompts, by a substantial margin, across all models. On GPT-3.5 for instance, \name achieves a win-rate of \final{80.6\%}, with an average of improvement of \final{4.9} points. \name also improves for open source models, with a win-rate of \final{91.2\%} on OPT-6.7 and an average improvement of \final{11.6} points. Finally, \name achieves similar gains on an instruction tuned model, with a win-rate of \final{89.1\%} and an average improvement of \final{7.3} points.

\paragraph{Performance when prompts are good}  We additionally investigate how \name's performance improvements change as a function of the performance of the base prompt. Hypothetically, one could imagine that better performing prompts are \textit{smoother} with respect to embeddings, thus diminishing (or negating) \name. In Figure \ref{fig:breakdown_and_smoothness} (upper left), we plot the improvement of \name against the performance of the base prompt for GPT-JT. Even when the base prompt performs well (i.e., F1 > $0.8$), \name improves on \final{89\%} of tasks by an average of \final{4.1} points.

\begin{table}[t!]
    \centering
    \begin{tabularx}{\textwidth}{llXX}
    & LM &  Win rate (\%)  & Avg. Improvement (F1) \\
    \toprule
    \multirow{2}{*}{API-Access Models} 
    & J1-Jumbo (176B) & \final{72.2} & \final{10.6} \\
    & GPT-3.5 ($>$ 170B) & \final{80.6} & \final{4.9} \\
    \midrule
    \multirow{2}{*}{Open Source} 
    & Bloom (7.1B) & \final{91.2} & \final{10.1} \\
    & OPT (6.7B) &  \final{91.2} & \final{11.6}\\
    \midrule
    \multirow{1}{*}{Instruction Tuned} 
    & GPT-JT (6B) &  \final{89.1} & \final{7.3}  \\
    \bottomrule
    \end{tabularx}
    \caption{We evaluate the extent to which \name improves the original prompt on different models in terms of win rate and relative improvement (defined in-line).  All models are run with three trials. For each model, we report the percentage of tasks (across all trials) for which \name improves, and the average improvement (in F1 points). Additional details provided in Appendix.}
    \label{tab:embroid_improvement}
\end{table}

\paragraph{Measuring performance in parameter count} A trend in recent literature has been to measure the magnitude of improvements to prompt performance in terms of parameter count~\cite{arora2022ask}, by showing how a particular method makes a smaller method equivalent in performance to a larger model. We find that \name enables the non-instructed tuned 1.3B GPT-Neo model to outperform an instruction tuned 6.7B model; across all trials, GPT-JT scores an average F1 of \final{67.8}, while \name+GPT-Neo-1.3B scores an average of \final{68.5}.

\subsection{Comparing prompt-patching to prompt-engineering}
Our work distinguishes between prompt-construction methods---which control how a prompt is generated---and prompt-patching methods---which attempt to identify and correct errors in the predictions produced by a prompt. We use \name to further study the difference between these frameworks in two ways. First, we compare \name's performance improvement over a base prompt to that of several specialized prompting strategies. Second, we examine the extent to which \name---when applied to the predictions produced by these prompting strategies---can generate further performance improvements. We study three prompting strategies: 
\begin{enumerate}
    \item Ensemble strategies, in which the predictions of multiple prompts are combined using an unsupervised ensembling model. Specifically, we compare to two ensembling methods previously studied for LLMs (AMA~\cite{arora2022ask} and majority vote~\cite{liu2023pre}), one ensembling method which incorporates embedding information (Liger~\cite{chen2022shoring}), and one well regarded weak supervision baseline (FlyingSquid~\cite{fu2020fast}). Each baseline is run over the predictions generated by three different prompts. 
    \item Chain-of-thought prompting~\cite{wei2022chain}, in which for each in-context demonstration, we provide a step-by-step explanation for the demonstration's label.
    \item Selective annotation (SA)~\cite{su2022selective}, in which we use embeddings to select a subset of $k$ data samples to label, and then, for each input sample, retrieve the most similar samples (under some embedding function) from this pool to use as in-context demonstrations.
\end{enumerate}

\paragraph{Ensemble methods} We evaluate two versions of \name. In the first version, we run \name with the predictions of only one prompt (\name-1). In the second version, we run \name with the predictions of three prompts (\name-3). The second version is comparable to applying \name to the outputs of an ensemble method. In Table \ref{tab:embroid_aggregation}, we observe that \name-1 is competitive with the ensemble baselines (while using substantially fewer predictions), while \name-3 consistently outperforms these baselines (across different LMs).

\begin{table}[t!]
    \centering
    \begin{tabularx}{\textwidth}{lcccccc}
    LM &  MV  & Liger & FlyingSquid & AMA & \name-1 & \name-3\\
    \toprule
     J1-Jumbo & \final{47.4} & \final{48.7} & \final{50.5}&\final{60.7}&\final{60.4}&\textbf{64.5}\\
     GPT-3.5& \final{81.4}& \final{82.5}&\final{82.1}&\final{84.7}&\final{83.9}&\textbf{86.0}\\
     Bloom-7.1B & \final{54.6}&\final{55.8}&\final{54.3}&\final{63.0}&\final{64.7}&\textbf{69.1}\\
     OPT-6.7 & \final{46.1}&\final{46.8}&\final{46.3}&\final{56.3}&\final{59.8}&\textbf{64.2}\\
     GPT-JT & \final{69.3}&\final{69.4}&\final{70.1}&\final{74.6}&\final{75.1}&\textbf{79.0}\\
    \bottomrule
    \end{tabularx}
    \caption{We evaluate how \name compares to common ensemble approaches for improving prompt prediction performance. All ensemble baselines are run with three sets of predictions. \name-1 is run with one set of predictions, and \name-3 is run with three set of predictions. For each method, we report the average macro-F1 over all tasks. We observe that \name-1 is competitive with ensemble methods which use many more predictions, while \name-3 outperforms all other methods by a substantial margin.}
    \label{tab:embroid_aggregation}
\end{table}

\paragraph{Chain-of-thought} We compare \name to chain-of-thought (CoT) prompting for a subset of a representative subset of 10 tasks on GPT-3.5. For each task, we manually construct a base prompt consisting of six demonstrations, and a CoT prompt where an explanation is provided for each demonstration. We first find that \name's performance improvement over the base prompt exceeds that of chain-of-thought prompting  (Table \ref{tab:cot}). Using \name to modify the base prompt is better on average than CoT prompting, outperforming CoT on six out of the ten tasks. Second, applying \name to predictions generated by a CoT prompt yields further improvements, outperforming vanilla CoT predictions on eight of ten tasks. 

\begin{figure*}
    \centering
    \begin{subfigure}[b]{0.45\textwidth}
        \centering
        \includegraphics[width=\textwidth]{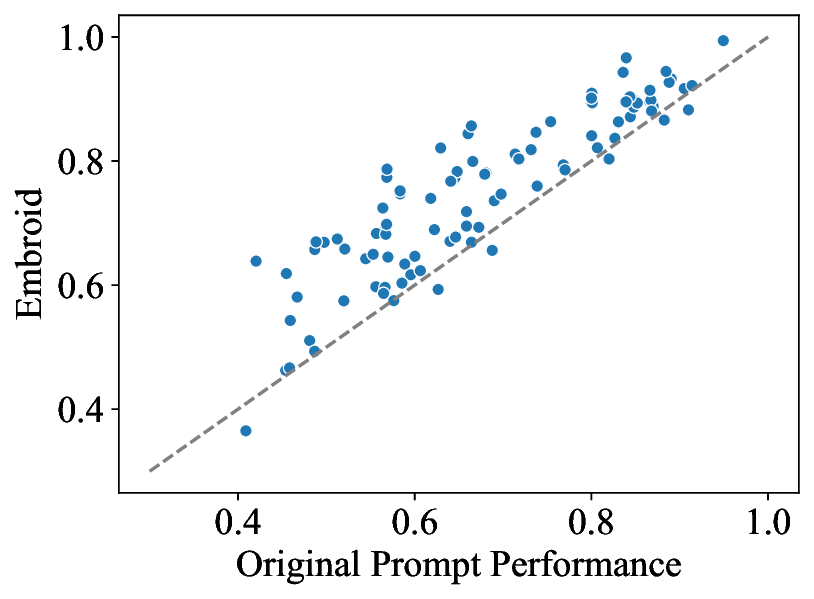}
    \end{subfigure}
    \begin{subfigure}[b]{0.45\textwidth}
        \centering
        \includegraphics[width=\textwidth]{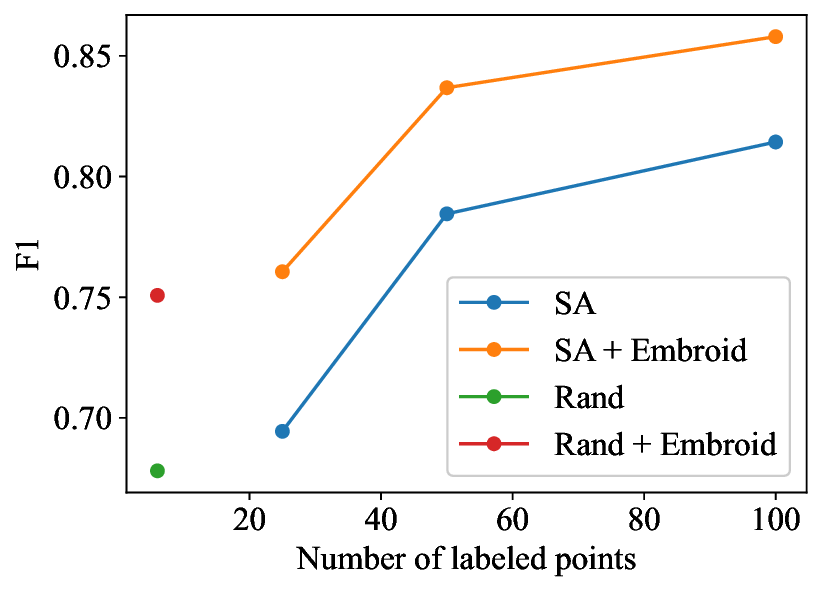}
    \end{subfigure}\\
    \begin{subfigure}[b]{0.45\textwidth}
        \centering
        \includegraphics[width=\textwidth]{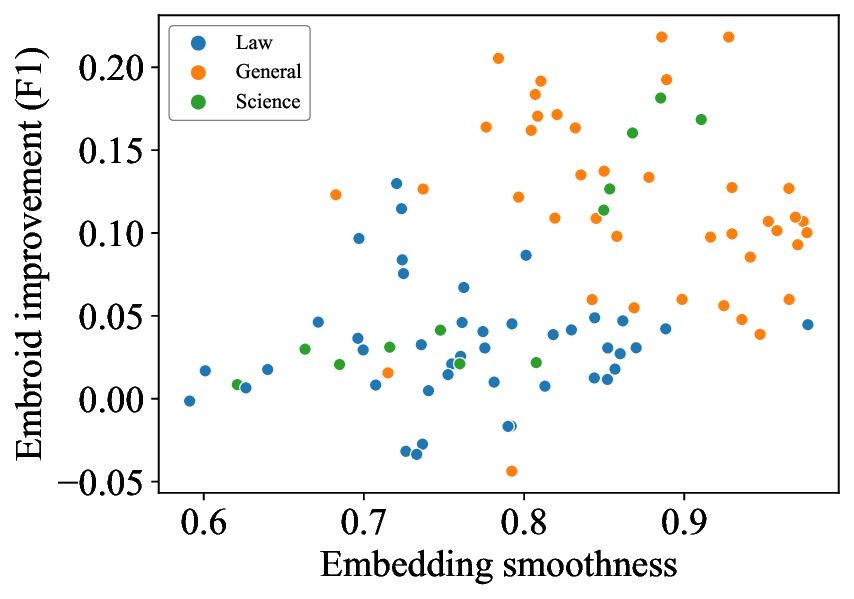}
    \end{subfigure}
    \begin{subfigure}[b]{0.45\textwidth}
        \centering
        \includegraphics[width=\textwidth]{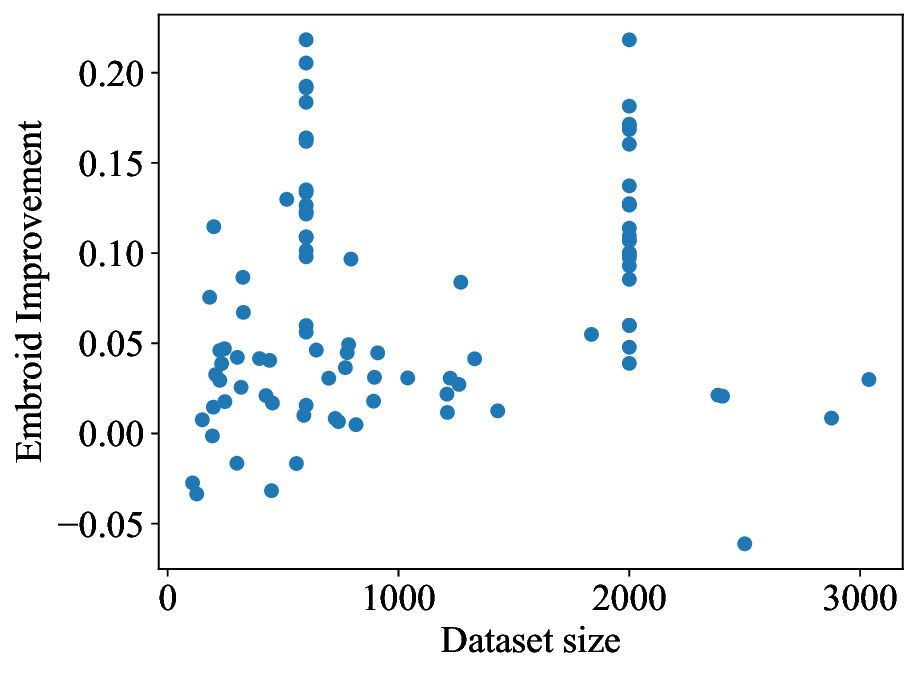}
    \end{subfigure}
    \caption{\textbf{Upper left}: The \name F1 plotted against the F1 score for the original prompt for GPT-JT. Even for high performing prompts, \name is capable of improving performance. The dashed line $y=x$ is plotted for visual aid. \textbf{Upper right}: A comparison of \name to selective annotation (SA) over all tasks for GPT-JT. \textbf{Bottom left}: For each task (using GPT-JT), we plot the performance improvement of \name against the average smoothness of the embeddings used. We observe a positive correlation ($r = 0.39$). \textbf{Bottom right}: Across all tasks, we measure the performance improvement of \name against the size of the task. }
    \label{fig:breakdown_and_smoothness}
\end{figure*}

\paragraph{Selective annotation (SA)} We compare \name to selective annotation with a label budget of $[6, 25, 50, 100]$ (Figure \ref{fig:breakdown_and_smoothness}, upper-right). For each task, we run selective annotation using a domain specific embedding. \name (applied to a prompt with randomly chosen samples) outperforms selective annotation with a label budget of 25 samples. When a label budget of 100 samples is available, \name improves the performance of a prompt constructed using selective annotation on 88\% of tasks, by an average of 4.3 points.

\begin{table}[t!]
    \centering
    \begin{tabularx}{\textwidth}{XXXX}
    Base prompt &  +CoT  & +\name & + CoT + \name \\
    \toprule
    76.3 & 80.1 & 81.9 & \textbf{85.4} \\
    \bottomrule
    \end{tabularx}
    \caption{We evaluate \name compared to, and applied to, CoT prompting on GPT-3.5 for a subset of 10 tasks. We report the average across the studied tasks.}
    \label{tab:cot}
\end{table}

\subsection{Ablations}
Finally, we perform several ablations of \name to study how performance changes as a function of (1) the domain specificity of the embedding used, (2) the quality of the embedding spaces used, and (3) the size of the dataset. Additional ablations are presented in Appendix \ref{sec:appendix:hyperparameters}.

\paragraph{Domain specific embeddings improve performance} We compare how performance on the legal and science tasks changes when we shift from domain specialized embeddings to general domain embeddings. On law tasks for GPT-JT, we find that using two legal embedding spaces outperforms using BERT and RoBERTa for \final{77\%} of tasks, by up to \final{6} points F1 on certain tasks~\cite{zheng2021does, henderson2022pile}. For science tasks for GPT-JT,  we find that using two science embedding spaces~\cite{lee2020biobert, beltagy2019scibert} outperforms using BERT and RoBERTa for \final{92\%} of tasks, by up to \final{4.3} points F1 on certain tasks.

\paragraph{Embedding quality} Building on Section \ref{sec:theory}, we compare \name's performance improvement over the base prompt to the average smoothness of the embedding spaces with respect to each task (Figure \ref{fig:breakdown_and_smoothness}). We observe a positive correlation: smoother embedding spaces are associated with larger performance gains (with a Pearson coefficient of $r = 0.39$). Applying this insight, we explore how performance changes when \textit{extremely} high quality embeddings are added. For a subset of 19 tasks we generate OpenAI \texttt{text-embedding-ada-002} embeddings, and find that adding them to \name improves performance by up to \final{13} points F1 (at an average of \final{2} points across all studied tasks).

\paragraph{Dataset size} Finally, we study how \name's performance improvement changes as the dataset size changes. Because \name relies on nearest-neighbors in different embedding spaces, we might expect performance to be poor when the dataset being annotated is small. In Figure \ref{fig:breakdown_and_smoothness} (bottom right), we see that \name achieves performance improvements even for ``small'' datasets with only several hundred samples. 
¯
\section{Conclusion}
We study the problem of improving prompt-based learning, by developing a method (\name) for detecting and correcting erroneous predictions without labeled data. We validate \name across a range of datasets and LMs, finding consistent improvement in many regimes. We take a moment to address the societal impact of our work: while we do not foresee any \textit{direct} harmful impacts arising from our work, we caution that any use of language models in meaningful applications should be accompanied by conversations regarding risks, benefits, stakeholder interests, and ethical safeguards.
\section{Acknowledgements}
We are grateful to Arjun Desai, Avanika Narayan, Benjamin Spector, Dilara Soylu, Gautam Machiraju, Karan Goel, Lucia Zheng, Rose E. Wang, Sabri Eyuboglu, Sarah Hooper, Simran Arora,  Maya Varma, and other members of the Hazy Research group for helpful feedback and conversations on this project.

We gratefully acknowledge the support of NIH under No. U54EB020405 (Mobilize), NSF under Nos. CCF1763315 (Beyond Sparsity), CCF1563078 (Volume to Velocity), 1937301 (RTML) and CCF2106707; US
DEVCOM ARL under No. W911NF-21-2-0251 (Interactive Human-AI Teaming); ONR under No.
N000141712266 (Unifying Weak Supervision); ONR N00014-20-1-2480: Understanding and Applying
Non-Euclidean Geometry in Machine Learning; N000142012275 (NEPTUNE); NXP, Xilinx, LETICEA, Intel, IBM, Microsoft, NEC, Toshiba, TSMC, ARM, Hitachi, BASF, Accenture, Ericsson,
Qualcomm, Analog Devices, Google Cloud, Salesforce, Total, the HAI-GCP Cloud Credits for
Research program, the Stanford Data Science Initiative (SDSI), the Wisconsin Alumni Research Foundation (WARF), the Center for Research on Foundation Models (CRFM), and members of the Stanford DAWN
project: Facebook, Google, and VMWare.
The U.S. Government is authorized to reproduce and distribute reprints for Governmental
purposes notwithstanding any copyright notation thereon.

Any opinions, findings, and conclusions or recommendations expressed in this material are those of the authors and do not necessarily reflect the views, policies, or endorsements, either expressed or implied, of NIH, ONR, or the U.S. Government.

\newpage
\bibliography{references}
\bibliographystyle{plain}


\newpage
\appendix
\onecolumn

\section{Notation}
The glossary is given in Table~\ref{table:glossary} below.

\begin{table*}[hp!]
\centering
\small
\begin{tabular}{l l}
\toprule
Symbol & Used for \\
\midrule
$x$ & Input sentence or paragraph $x \in \X$. \\
$y$ & Binary task label $y \in \Y = \{-1, +1\}$. \\
$\data$ & Unlabeled dataset $\data = \{x_i\}_{i = 1}^{n_u}$ of $n_u$ points. \\
$\prob, \prob_x$ & The joint distribution of $(x, y)$ and the marginal on $x$, respectively. \\
$n_l$ & Number of labeled in-context examples used in querying the LLM (5-10 examples).  \\
$\llm(\cdot), \phi(\cdot)$ & Users interact with a language model $\llm$ via a prompt $\phi$ on $x$. \\
$\nlf$ & Number of prompts that we have access to. \\
$\lambdab$ & $\lambdab(x) = [\lambda_1(x), \dots, \lambda_{\nlf}(x)]$ where $\lambda_i(x)$ is shorthand for $\llm(\phi_i(x))$. \\
$\EmbS$ & The set of $N$ embedding models, $\EmbS = \{E_1, \dots, E_N\}$ where each embedding is represented as a \\
& fixed mapping $E_i:\X \mapsto \Z_i$. \\
$Z$ & Partition function for normalization of~\eqref{eq:base}. \\
$\theta_y, \theta$ & $\theta_y$ is a label balance parameter and each $\theta_i$ is a scalar accuracy parameter for the $i$th source in~\eqref{eq:base}. \\
$\nnx_{j,k}(x)$ & The $k$-nearest neighbors of $x$ in embedding space $\emb_j$. \\
$\svotes$ & $\svotes(x) = [\svote{1}, \dots, \svote{\nemb}] \in \{-1, 0, 1\}^{\nemb \nlf}$, where $\svote{j} = [\svote{j}[1], \dots \svote{j}[\nlf]]$ and $\svote{j}[i](x)$ is the \\
& smoothed neighborhood prediction of $\lambda_i(x)$ in $\emb_j$ (eq.~\eqref{eq:svote}). \\
$\shrink^{+}_i$,$\shrink^{-}_i$ & Shrinkage parameters for determining when $\svote{j}[i](x)$ is set to $0$, $-1$, or $1$. \\
$\alpha_j$ & Vector of $m$ accuracy parameters for $\emb_j$ when used with $m$ prompts in~\eqref{eq:main}. \\ 
$L(\lambdab, \svote, \data)$ & Generalization error of \name ( expected cross-entropy loss). \\
$a_{\max}$ & The largest scaled accuracy of any source, $a_{\max} = \max_i \E{}{\lambda_i(x) y}$. \\
$b_{\min}$ & The smallest expected pairwise product between any two sources, $b_{\min} = \min_{i,j}\{\E{}{\lambda_i \lambda_j}, \Ehat{\lambda_i \lambda_j}\}$. \\
$M_E(\cdot)$ & An embedding is $M$-smooth if $\Pr(\tilde{y} = c | y = c, \|E(x) - E(\tilde{x}) \| \le \varepsilon) \ge M_E(\varepsilon)$ for all $c \in \Y$ and any $\varepsilon > 0$, \\
& where $M_E(\cdot) \in [0, 1]$ is decreasing in its input. \\
$\beta_i$ & The accuracy of $\lambda_i$, $\beta_i = \Pr(\lambda_i = y)$. \\
$p_{\lambdab}$ & The prediction on $x_0$ given only access to $\lambdab$, $p_{\lambdab} = \Pr(y = 1 | \lambdab(x_0))$. \\
$\varepsilon_k$ & The maximum distance between $x_0$ and its $k$ neighbors, $\varepsilon_k = \max_{\tilde{x} \in \nnx_k(x)} \|E(x) - E(\tilde{x})\|$. \\
\toprule
\end{tabular}
\caption{
	Glossary of variables and symbols used in this paper.
}
\label{table:glossary}
\end{table*}

\newpage

\newpage
\section{Weak supervision background}\label{app:wsback}
In this section, we provide details on the inference and learning procedures for solving the graphical model defined in equation~\eqref{eq:base}. The content from this section is derived from~\cite{fu2020fast} and~\cite{chen2022shoring}.

\paragraph{Pseudolabel inference} To perform inference, we compute $\hat{\Pr}(y | \lambdab(x))$ for some $x\in\X$.  This is done via Bayes' rule and the conditional independence of weak sources: 
\begin{align} \label{eq:bayes}
    \Pr(y = 1 | \lambdab(x)) = \frac{\prod_{i = 1}^m \Pr(\lambda_i(x) | y = 1) \Pr(y = 1)}{\Pr(\lambdab(x))}. 
\end{align}

We assume that the class balance is known; for our datasets, the class balance is $\Pr(y = 1) = 0.5$.  More generally, it can be estimated~\cite{ratner19}. The latent parameter of interest in this decomposition is $\Pr(\lambda_i = 1 | y = 1)$, which corresponds to the accuracy of $\lambda_i$.

\begin{algorithm}[h!]
\caption{Triplet method~\cite{fu2020fast}}
\begin{algorithmic}
    \STATE \textbf{Input:}
    Dataset $\data$, weak sources $\lambdab(x)$.
    \FOR{$i \in [m] $}
        \FOR{$j, k \in [m] \backslash i$}
            \STATE Estimate $\Ehat{\lambda_i \lambda_j}$ over $\data$, and similarly estimate $\Ehat{\lambda_i \lambda_k }$ and $\Ehat{\lambda_j \lambda_k}$.
            \STATE Compute $\hat{a}_i^{j, k} = \sqrt{\bigg|\frac{\Ehat{\lambda_i \lambda_j} \Ehat{\lambda_i \lambda_k }}{\Ehat{\lambda_j \lambda_k}} \bigg|}$.  
        \ENDFOR
        \STATE Calculate average $\hat{a}_i = \text{Mean}(\hat{a_i}^{j, k} \;\; \forall j, k \in [m] \backslash i)$.
        \STATE Compute estimated accuracy $\hat{\Pr}(\lambda_i = y) = \frac{1 + \hat{a}_i}{2}$.
    \ENDFOR 
    \STATE {\bfseries Output:} Accuracies $\hat{\Pr}(\lambda_i = y)$ for all $i \in [m]$.
\end{algorithmic}
\label{alg:triplet}
\end{algorithm}

\paragraph{Source parameter estimation: Triplet method} Previous approaches have considered how to estimate $\Pr(\lambda_i = 1 | y = 1)$ via the \emph{triplet method}~\cite{fu2020fast}, which exploits conditional independence properties.
First, by the properties of the graphical model in~\eqref{eq:base}, it holds that the accuracy of $\lambda_i$ is symmetric: $\Pr(\lambda_i = 1 | y = 1) = \Pr(\lambda_i = -1 | y = -1) = \Pr(\lambda_i = y)$ (Lemma $4$ of~\cite{chen2022shoring}). Therefore, $\Pr(\lambda_i = 1 | y = 1)$ can be written in terms of $\E{}{\lambda_i y}$ with $\E{}{\lambda_i y} = 2\Pr(\lambda_i = 1 | y = 1) - 1$. 

Define $a_i = \E{}{\lambda_i y}$. The graphical model in~\eqref{eq:base} tells us that $\lambda_i y \independent \lambda_j y$ if $\lambda_i \independent \lambda_j | y$, which holds for all $i, j \in [\nlf]$ (Proposition $1$ of~\cite{fu2020fast}).  As a result, $\E{}{\lambda_i y} \times \E{}{\lambda_j y} = \E{}{\lambda_i \lambda_j y^2} = \E{}{\lambda_i \lambda_j}$, which is a quantity that can be computed from observed LLM predictions. That is, we have that $a_i a_j = \E{}{\lambda_i \lambda_j}$. If we introduce a third $\lambda_k$, we can generate a system of equations over $a_i, a_j, a_k$ in terms of their pairwise rates of agreements:
\begin{align}
    a_i a_j &= \E{}{\lambda_i \lambda_j} \\
    a_i a_k &= \E{}{\lambda_i \lambda_k} \\
    a_j a_k &= \E{}{\lambda_j \lambda_k}.
\end{align}

Solving, we get that
\begin{align}
    |a_i| := \sqrt{\bigg| \frac{\E{}{\lambda_i \lambda_j} \E{}{\lambda_i \lambda_k }}{\E{}{\lambda_j \lambda_k }}\bigg|},
\end{align}

and likewise for $a_j, a_k$. If we assume that each weak source is better than random over the dataset, then $a_i = |a_i| > 0$, so we can uniquely recover the accuracy of each source by selecting two other sources and computing the above expression by using empirical expectations over $\data$. We then set $\hat{\Pr}(\lambda_i = 1 | y = 1) = \frac{1 + \hat{a}_i}{2}$ and plug this into the expression for $\Pr(y  = 1 | \lambdab(x))$ in~\eqref{eq:bayes}.

This approach is formally described in Algorithm~\ref{alg:triplet}.

\newpage
\section{Proofs} \label{appendix:proofs}

\subsection{Proof of proposition~\ref{prop:gen_err}}
We note that $[\lambdab, \svotes]$ can be viewed as a set of sources in the weak supervision set up used in~\cite{fu2020fast, chen2022shoring}. Therefore, we can apply Theorem $1$ from~\cite{chen2022shoring} to our problem setting, noting that we do not perform their clustering step and that our predictions do not abstain and output $0$ in addition to $\{-1, 1\}$. We have a total of $(\nemb + 1)\nlf$ sources, so
\begin{align}
    L(\lambdab, \svotes, \data) &\le H(y | \lambdab, \svotes) +  \frac{3(1 - b_{\min}^2)}{8b_{\min}^2 (1 - a_{\max}^2)} \bigg(\frac{1}{b_{\min}^4} + \frac{2}{b_{\min}^2} \bigg) \frac{(\nemb + 1)\nlf}{\nul} +  o(1/\nul).
\end{align}

\subsection{Proof of theorem~\ref{thm:cond_entr_smoothness}}

We can write the change in point-wise irreducible error as follows:
\begin{align}
    H(y | \lambdab(x_0)) - H(y | \lambdab(x_0), \svotes(x_0)) &= \E{}{- \log \Pr(y | \lambdab(x_0))  + \log \Pr(y | \lambdab(x_0), \svotes(x_0))} \\
    &= \E{}{\log \frac{\Pr(y | \lambdab(x_0), \svotes(x_0))}{\Pr(y | \lambdab(x_0))}} \\
    &= \E{}{\log \bigg(\frac{\Pr(\lambdab(x_0), \svotes(x_0) | y) \Pr(y)}{\Pr(\lambdab(x_0), \svotes(x_0))} \cdot \frac{\Pr(\lambdab(x_0))}{\Pr(\lambdab(x_0)| y) \Pr(y)} \bigg)} \\
    &= \E{}{\log \frac{\Pr(\svotes(x_0) | \lambdab(x_0), y)}{\Pr(\svotes(x_0) | \lambdab(x_0))}}.
\end{align}

Next, we use the fact that $\lambdab(x_0) \independent \svotes(x_0) | y$ to simplify the expression into
\begin{align}
    \E{}{\log \frac{\Pr(\svotes(x_0) | y)}{\Pr(\svotes(x_0) | y = 1) \Pr(y = 1 | \lambdab(x_0)) + \Pr(\svotes(x_0) | y = -1) \Pr(y = -1 | \lambdab(x_0))}}.
\end{align}

The exact $\svotes(x_0)$ is unknown but is drawn from the distribution $\Pr(\svotes | y = 1)$ since $x_0$'s label is $1$.
 Then, this expression becomes an expectation over $\svotes$:
\begin{align}
    \E{\svotes | y = 1}{\log \frac{\Pr(\svotes | y = 1)}{\Pr(\svotes | y = 1) p_{\lambdab} + \Pr(\svotes | y = -1) (1 - p_{\lambdab})}}.
\end{align}

Given that our $\svotes$ is high-quality, we suppose that $\svotes(x_0)$ all equal $1$ with high probability, and then we can lower bound our expression by
\begin{align}
    \Pr(\svotes = 1 | y = 1) \log \frac{\Pr(\svotes = 1 | y = 1)}{\Pr(\svotes = 1 | y = 1) p_{\lambdab} + \Pr(\svotes = 1 | y = -1) (1 - p_{\lambdab})}. \label{eq:p_log_gain}
\end{align}

The key quantity of interest is $\Pr(\svotes = 1 | y = 1) = \prod_{i = 1}^m \Pr(\svote{}[i] = 1 | y = 1)$. We focus on bounding $\Pr(\svote{}[i] = 1 | y = 1)$ next. Suppose that the $k$ neighbors of $x_0$ are $x_1, \dots, x_k$. Define $p_j = \Pr(\lambda_i(x_j) = 1 | y = 1)$ for all $j \in [k]$. Note that $\lambda_i(x_j) \independent \lambda_i(x_{j'}) | y$ for any $j, j' \in [k]$ (while $\svote{}[i]$ as a whole is dependent on $y$, individual neighbors are still conditionally independent). Then, the event that $\svote{}[i] = 1 | y = 1$ is as least as likely as the event that $\text{Binomial}(k, \min_{i \in [k]} p_i) \ge \frac{k}{2}$. Let $p_{\min} = \min_{i \in [k]} p_i$, and assume that $p_{\min} \ge \frac{1}{2}$. Then,
\begin{align}
    \Pr(\svote{}[i] = 1 | y = 1) &\ge \Pr\Big(\text{Binomial}(k, p_{\min}) \ge \frac{k}{2}\Big) = \Pr\Big(\frac{1}{k} \sum_{j = 1}^k X_j \ge \frac{1}{2}\Big) \\
    &= \Pr\Big(\frac{1}{k} \sum_{j = 1}^k X_j \ge p_{\min} - \Big(p_{\min} - \frac{1}{2}\Big)\Big),
\end{align}

where $X_j \sim \text{Bernoulli}(p_{\min})$. Next, let $\delta = p_{\min} - \frac{1}{2}$. We can apply Hoeffding's inequality to get
\begin{align}
    \Pr\Big(\text{Binomial}(k, p_{\min}) \ge \frac{k}{2}\Big) &= \Pr\Big(\frac{1}{k} \sum_{j = 1}^k X_j \ge p_{\min} - \delta \Big) = 1 - \Pr\Big(\frac{1}{k} \sum_{j = 1}^k X_j \le p_{\min} - \delta \Big) \ge 1 - \exp(- 2 \delta^2 k) \\
    &= 1 - \exp(-2k (p_{\min} - 0.5 )^2 ).
\end{align}

All that's left is to lower bound $p_{\min}$. Without loss of generality, suppose that $p_{\min}$ corresponds to an arbitrary $p_j = \Pr(\lambda_i(x_j)  = 1 | y = 1)$. We can decompose this probability into 
\begin{align}
    \Pr(\lambda_i(x_j)  &= 1 | y = 1) = \Pr(\lambda_i(x_j)  = 1, y(x_j) = 1 | y = 1) + \Pr(\lambda_i(x_j) = 1, y(x_j) = -1 | y = 1) \\
    &= \Pr(\lambda_i(x_j) = 1 | y(x_j) = 1) \Pr(y(x_j) = 1 | y = 1) + \Pr(\lambda_i(x_j) = 1 | y(x_j) = -1) \Pr(y(x_j) = -1 | y = 1).
\end{align}

Since $\Pr(\lambda_i(x_j) = 1 | y(x_j) = 1)$ is over all $x_j \sim \prob_x$, this quantity is just equal to the accuracy of $\lambda_i$, $a_i$. Next, recall that $\| \emb(x_j) - \emb(x)\| \le \varepsilon_k$, where $\varepsilon_k = \max_{x_i \in \nnx(x)} \|\emb(x) - \emb(x_i)\|$ is the maximum distance from the $k$ neighbors to $x$. Then, we can write $\Pr(y(x_j) = 1 | y = 1)$ as $\Pr(y(x_j) = 1 | y = 1, \|\emb(x_j) - \emb(x)\| \le \varepsilon_k) \ge M_{\emb}(\varepsilon_k)$, since we have assumed that $E$ is $M$-smooth. We can now bound $p_{\min}$:
\begin{align}
    p_{\min} \ge a_i M_{\emb}(\varepsilon_k) + (1 - a_i) (1 - M_{\emb}(\varepsilon_k)).
\end{align}

Therefore, we have that
\begin{align}
    \Pr(\svotes = 1 | y = 1) \ge \prod_{i = 1}^m \big[1 - \exp[-2k( a_i M_{\emb}(\varepsilon_k) - 0.5)^2]\big].
\end{align}

Before we plug in $\Pr(\svotes = 1 | y = 1)$ into~\eqref{eq:p_log_gain}, we simplify the expression. Note that $\Pr(\svotes = 1 | y = 1)$ can be written as $\prod_{i = 1}^m p_i$ for some $p_i$, and $\Pr(\svotes = 1 | y = -1)$ can be written as $\prod_{i = 1}^m (1 - p_i)$. A simple proof by induction shows that $\prod_{i = 1}^m (1 - p_i) \le 1 - \prod_{i = 1}^m p_i$. Therefore, we can write that~\eqref{eq:p_log_gain} is lower bounded by
\begin{align}
    \Pr(\svotes = 1 | y = 1) \log \frac{\Pr(\svotes = 1 | y = 1)}{\Pr(\svotes = 1 | y = 1) p_{\lambdab} + (1 - \Pr(\svotes = 1 | y = 1))(1 - p_{\lambdab})}
\end{align}

Let's abbreviate $\Pr(\svotes = 1 | y = 1)$ as $x$ and define the function
\begin{align}
    f(x) = x \log \frac{x}{x p_{\lambdab} + (1 - x) (1 - p_{\lambdab})}.
\end{align}

We note that for $x \ge 0.5$, $f(x)$ is convex and can thus be lower bounded by $f(x) \ge f'(0.5)(x - 0.5)$. We compute $f'(x) = \frac{1 - p_{\lambdab}}{x p_{\lambdab} + (1 - x) (1 - p_{\lambdab})}$, so $f'(0.5) = 2(1 - p_{\lambdab})$. Therefore, $f(x) \ge 2(1 - p_{\lambdab})(x - 0.5)$. Our final bound on the pointwise difference in irreducible error on $x_0$ is
\begin{align}
    H(y | \lambdab(x_0)) - H(y | \lambdab(x_0), \svotes(x_0)) \ge 2(1 - p_{\lambdab}) \bigg[ \prod_{i = 1}^m \big[1 - \exp[-2k( a_i M_{\emb}(\varepsilon_k) - 0.5)^2]\big] - 0.5 \bigg] .
\end{align}

\paragraph{Information gain from using $\lambdab, \svotes$ over $\svotes$} We briefly comment on the opposite direction---how much does using both LLM predictions and neighborhood predictions help over just using neighborhood predictions? 

The quantity we aim to lower bound is $H(y | \svotes(x_0)) - H(y | \lambdab(x_0), \svotes(x_0))$ for a point of interest $x_0$. We can write this quantity as 
\begin{align}
    H(y | \svotes(x_0)) - H(y | \lambdab(x_0), \svotes(x_0)) = \E{}{\log \frac{\Pr(\lambdab(x_0) | y)}{\Pr(\lambdab(x_0) | \svotes(x_0))}}
\end{align}

Without loss of generality, suppose that the true label on $x_0$ is $y = 1$, and that for each $\lambda_i$, the neighborhood around $x_0$ consists of a balanced mix of $\lambda_i = 1$ and $\lambda_i = -1$. Then, with high probability we have that $\Pr(y = 1 | \svotes(x_0)) = p_{\svotes} \approx 0.5$. From our proof of Theorem~\ref{thm:cond_entr_smoothness}, we can thus write
\begin{align}
    &H(y | \svotes(x_0)) - H(y | \lambdab(x_0), \svotes(x_0)) = \E{\lambda(x_0)}{\log \frac{\Pr(\lambdab(x_0) | y = 1)}{\Pr(\lambda(x_0) | y = 1) p_{\svotes} + \Pr(\lambda(x_0) | y = -1) (1 - p_{\svotes})}} \\
    &\ge \Pr(\lambdab(x_0) = 1 | y = 1))\log \frac{\Pr(\lambdab(x_0) = 1 | y = 1)}{\Pr(\lambdab(x_0) = 1 | y = 1) p_{\svotes} + \Pr(\lambdab(x_0) = 1 | y = -1) (1 - p_{\svotes})} \\
    &\ge 2(1 - p_{\svotes}) (\Pr(\lambdab(x_0) = 1 | y = 1) - 0.5)
\end{align}

If $\lambdab$ on $x_0$ has high accuracy and $p_{\svotes}$ is low, then we can have significant point-wise information gain from modeling both $\lambdab$ and $\svotes$ rather than just $\svotes$.

\newpage
\section{Datasets}\label{appendix:datasets}

\paragraph{Motivation} We study the performance of our method across a diverse collection of \textit{task definitions}. In our setting, a task definition denotes a specific classification that a data scientist wishes to perform. For instance, a data scientist working on quantifying the breadth of legal issues that individuals face may wish to identify which posts in an online forum refer implicate legal issues related to housing.  

This evaluation strategy is motivated by the observation that task definitions vary in their smoothness across embedding spaces, as different embeddings may do a better job of capturing features relevant for the task. For instance, out-of-the-box Sentence-BERT embeddings are better than traditional BERT at capturing the topicality of a sentence~\cite{reimers2019sentence}. By focusing on a broad range of task definitions, we can better forecast how our method might perform for new tasks that practitioners may need to create classifiers for. We also avoid issues with leakage that may arise as the practice of finetuning LLMs on tasks increases~\cite{chung2022scaling}.

In total, we study \ntasks~distinct task definitions, encompassing 100,418 total samples. Each task varies between 108 and 3308 samples. 

\paragraph{Legal tasks} The emergence of LLMs is exciting for law and finance, where expert-annotations are especially difficult to acquire~\cite{guha2022legalbench, bommasani2021opportunities}. Drawing on recent benchmarks and released datasets framing the potential use cases for LLMs in law, we study the following datasets:
\begin{itemize}
    \item CUAD~\cite{cuad}: The original CUAD dataset consists of 500 contracts spanning an array of sectors, with clauses manually into one of 41 legal categories. Following~\cite{guha2022legalbench}, we adapt the original dataset for clause-by-clause classification. We turn each clause type into a binary classification task, where the objective is to distinguish clauses of that type from clauses of other types (i.e. ``negatives''). Negative clauses are sampled randomly so as to make the task class balanced. We ignore clauses for which there are insufficient annotations in the original dataset.
    \item Learned Hands~\cite{learnedhands}: The Learned Hands dataset consists of legal questions that individuals publicly posted to an online forum (r/legaladvice). The questions have been coded by experts into legal categories according to the Legal Issues Taxonomy~\cite{list}. We consider several such issue classes, and create a binary classification task for each issue. Negative clauses are sampled randomly so as to make the task class balanced. Because these questions can be long, we truncate them at 50 tokens.
\end{itemize}

\paragraph{Science tasks} LLMs have generated similar excitement for medical and science informatics applications~\cite{bommasani2021opportunities, agrawal2022large}. We study established classification/extraction benchmarks.

\begin{itemize}
    \item Chemprot~\cite{chemprot}: ChemProt consists of sentences from PubMed abstracts describing chemical-protein relationships. We study seven relations, and create a binary task for each one. Each task is class balanced, with negatives sampled from the other relations.
    \item RCT~\cite{rct}: The RCT dataset consists of PubMed abstracts for papers discussing randomized control trials, where sentences in the abstract are annotated according to their semantic role (e.g., background, methods, results, etc). There are five roles, and we create a binary task for each one. Each task is class balanced, with negatives sampled from the other relations.
\end{itemize}

\paragraph{General domain tasks} Finally, we study the following ``general domain'' tasks, derived from popular sentence classification and information extraction benchmarks.

\begin{itemize}
    \item FewRel~\cite{fewrel}:  This is a relationship classification/extraction dataset, where each sample corresponds to a sentence mentioning the relationship between two entities. We select 20 relations, and for each relation construct a binary classification task with 700 positive instances of the relation, and 700 randomly sampled sentences (corresponding to other relations).
    \item Spam Detection~\cite{wrench}: We study the YouTube  spam detection task from the WRENCH benchmark. This task requires classifying YouTube comments as spam/not spam.
    \item Toxic content detection~\cite{civilcomments}: This task requires classifying whether posted comments are toxic or not. We use a sampled subset of the CivilComments dataset. 
    \item AG News~\cite{zhang2015character}: The original dataset organizes news snippets into four categories: World, Sports, Business, and Science/Technology. We create a separate task for each category. Negatives are sampled from the remaining classes. 
    \item DBPedia~\cite{zhang2015character}: DBPedia is a 14-way ontology classification dataset. We convert this into 14 distinct tasks, corresponding to each of the ontology types. 
\end{itemize}

{\small\begin{xltabular}{\textwidth}{lXc}
\caption{Legal tasks} \label{tab:legal_tasks} \\

\multicolumn{1}{l}{\textbf{Task}} & \multicolumn{1}{l}{\textbf{Description/Intent}} & \multicolumn{1}{l}{\textbf{Size}} \\ \toprule 
\endfirsthead

\multicolumn{3}{c}%
{\tablename\ \thetable{} -- continued from previous page} \\
\multicolumn{1}{l}{\textbf{Task}} & \multicolumn{1}{l}{\textbf{Description/Intent}} & \multicolumn{1}{l}{\textbf{Size}} \\ \toprule
\endhead

\endfoot

\bottomrule
\endlastfoot
Affiliate License-Licensee (CUAD) & Does the clause describe a license grant to a licensee (incl. sublicensor) and the affiliates of such licensee/sublicensor? & 208 \\ \midrule
Anti-Assignment (CUAD) & Does the clause require consent or notice of a party if the contract is assigned to a third party? & 1212 \\ \midrule
Audit Rights (CUAD) & Does the clause discuss potential audits? & 1224 \\ \midrule
Cap On Liability (CUAD) & Does the clause specify a cap on liability upon the breach of a party’s obligation? & 1262 \\ \midrule
Change Of Control (CUAD) & Does the clause give one party the right to terminate if such party undergoes a change of control? & 426 \\ \midrule
Competitive Restriction Exception (CUAD) & Does the clause mention exceptions or carveouts to Non-Compete, Exclusivity and No-Solicit of Customers? & 226 \\ \midrule
Covenant Not To Sue (CUAD) & Does the clause mention if a party is restricted from contesting the validity of the counterparty’s ownership of intellectual property? & 318 \\ \midrule
Effective Date (CUAD) & Does the clause mention when the contract becomes effective? & 246 \\ \midrule
Exclusivity (CUAD) & Does the clause mention an exclusive dealing commitment with the counterparty? & 770 \\ \midrule
Expiration Date (CUAD) & Does the clause mentions a date when the contract's term expires? & 892 \\ \midrule
Governing Law (CUAD) & Does the clause mentions which state/country's laws govern interpretation of the contract? & 910 \\ \midrule
Insurance (CUAD) & Does the clause mention a requirement for insurance? & 1040 \\ \midrule
Ip Ownership Assignment (CUAD) & Does the clause mention if intellectual property created by one party become the property of the counterparty? & 590 \\ \midrule
Irrevocable Or Perpetual License (CUAD) & Does the clause describe a license grant that is irrevocable or perpetual? & 300 \\ \midrule
Joint Ip Ownership (CUAD) & Does the clause provide for joint or shared ownership of intellectual property between the parties to the contract? & 198 \\ \midrule
License Grant (CUAD) & Does the clause describe a license granted by one party to its counterparty? & 1430 \\ \midrule
Liquidated Damages (CUAD) & Does the clause award either party liquidated damages for breach or a fee upon the termination of a contract (termination fee)? & 226 \\ \midrule
Minimum Commitment (CUAD) & Does the clause specifies a minimum order size or minimum amount or units pertime period that one party must buy from the counterparty? & 778 \\ \midrule
No-Solicit Of Employees (CUAD) & Does the clause restricts a party’s soliciting or hiring employees and/or contractors from the counterparty, whether during the contract or after the contract ends (or both). & 150 \\ \midrule
Non-Compete (CUAD) & Does the clause restrict the ability of a party to compete with the counterparty or operate in a certain geography or business or technology sector? & 450 \\ \midrule
Non-Disparagement (CUAD) & Does the clause require a party not to disparage the counterparty? & 108 \\ \midrule
Non-Transferable License (CUAD) & Does the clause limit the ability of a party to transfer the license being granted to a third party? & 558 \\ \midrule
Notice Period To Terminate Renewal (CUAD) & Does the clause requires a notice period to terminate renewal? & 234 \\ \midrule
Post-Termination Services (CUAD) & Does the clause subject a party to obligations after the termination or expiration of a contract, including any post-termination transition, payment, transfer of IP, wind-down, last-buy, or similar commitments? & 816 \\ \midrule
Renewal Term (CUAD) & Does the clause mention a renewal term for after the initial term expires? & 398 \\ \midrule
Revenue-Profit Sharing (CUAD) & Does the clause require a party to share revenue or profit with the counterparty for any technology, goods, or services? & 784 \\ \midrule
Rofr-Rofo-Rofn (CUAD) & Does the clause provide a party with a right of first refusal? & 698 \\ \midrule
Source Code Escrow (CUAD) & Does the clause requires one party to deposit its source code into escrow with a third party, which can be released to the counterparty upon the occurrence of certain events (bankruptcy, insolvency, etc.)? & 126 \\ \midrule
Termination For Convenience (CUAD) & Does the clause state that one party can terminate this contract without cause (solely by giving a notice and allowing a waiting period to expire)? & 442 \\ \midrule
Uncapped Liability (CUAD) & Does the clause state that a party’s liability is uncapped upon the breach of its obligation in the contract? & 302 \\ \midrule
Volume Restriction (CUAD) & Does the clause describe a fee increase or consent requirement if one party’s use of the product/services exceeds certain threshold? & 328 \\ \midrule
Warranty Duration (CUAD) & Does the clause mentions the duration of any warranty against defects or errors in technology, products, or services provided under the contract? & 326 \\ \midrule
BU (Learned Hands) & Does the text discuss issues relating to business or intellectual property? & 200 \\ \midrule
CO (Learned Hands) & Does the text discuss issues relating to courts and lawyers? & 194 \\ \midrule
CR (Learned Hands) & Does the text discuss issues relating to criminal issues? & 644 \\ \midrule
ES (Learned Hands) & Does the text discuss issues relating to estates or wills? & 182 \\ \midrule
FA (Learned Hands) & Does the text discuss issues relating to family or divorce? & 794 \\ \midrule
HE (Learned Hands) & Does the text discuss issues relating to health? & 248 \\ \midrule
HO (Learned Hands) & Does the text discuss issues relating to housing? & 1270 \\ \midrule
MO (Learned Hands) & Does the text discuss issues relating to payments or debt? & 740 \\ \midrule
TO (Learned Hands) & Does the text discuss issues relating to accidents or harassment? & 454 \\ \midrule
TR (Learned Hands) & Does the text discuss issues relating to cars or traffic? & 516 \\ \midrule
WO (Learned Hands) & Does the text discuss issues relating to employment or job? & 726 \\
\end{xltabular}
}
{\small
\begin{xltabular}{\textwidth}{lXc}
\caption{Science tasks} \label{tab:science_tasks} \\

\multicolumn{1}{l}{\textbf{Task}} & \multicolumn{1}{l}{\textbf{Description/Intent}} & \multicolumn{1}{l}{\textbf{Size}} \\ \toprule 
\endfirsthead

\multicolumn{3}{c}%
{\tablename\ \thetable{} -- continued from previous page} \\
\multicolumn{1}{l}{\textbf{Task}} & \multicolumn{1}{l}{\textbf{Description/Intent}} & \multicolumn{1}{l}{\textbf{Size}} \\ \toprule
\endhead

\endfoot

\bottomrule
\endlastfoot
Agonist (Chemprot) & Does the sentence describe an agonist relationship? & 896 \\ \midrule
Antagonist (Chemprot) & Does the sentence describe an antagonist relationship? & 1330 \\ \midrule
Downregulator (Chemprot) & Does the sentence describe a downregulator relationship? & 3038 \\ \midrule
Part\_of (Chemprot) & Does the sentence describe a part-of relationship? & 1210 \\ \midrule
Regulator (Chemprot) & Does the sentence describe a regulator relationship? & 2876 \\ \midrule
Substrate (Chemprot) & Does the sentence describe a substrate relationship? & 2384 \\ \midrule
Upregulator (Chemprot) & Does the sentence describe an upregulator relationship? & 2404 \\ \midrule
Background (RCT) & Does the sentence describe background on the study? & 2000 \\ \midrule
Conclusions (RCT) & Does the sentence state a conclusion? & 2000 \\ \midrule
Methods (RCT) & Does the sentence describe a scientific experimental method? & 2000 \\ \midrule
Objective (RCT) & Does the sentence describe the goal of the study? & 2000 \\ \midrule
Results (RCT) & Does the sentence describe experimental results? & 2000 \\ 
\end{xltabular}
}
{\small
\begin{xltabular}{\textwidth}{lXc}
\caption{General domain tasks} \label{tab:general_domain_tasks} \\

\multicolumn{1}{l}{\textbf{Task}} & \multicolumn{1}{l}{\textbf{Description/Intent}} & \multicolumn{1}{l}{\textbf{Size}} \\ \toprule 
\endfirsthead

\multicolumn{3}{c}%
{\tablename\ \thetable{} -- continued from previous page} \\
\multicolumn{1}{l}{\textbf{Task}} & \multicolumn{1}{l}{\textbf{Description/Intent}} & \multicolumn{1}{l}{\textbf{Size}} \\ \toprule
\endhead

\endfoot

\bottomrule
\endlastfoot 
Business (AGNews) & Does the article discuss business news? & 2000 \\ \midrule
Sports (AGNews) & Does the article discuss sports news? & 2000 \\ \midrule
Technology (AGNews) & Does the article discuss technology news? & 2000 \\ \midrule
World (AGNews) & Does the article discuss global affairs or world events? & 2000 \\ \midrule
Civil Comments & Does the sentence contain toxic or hateful content? & 2500 \\ \midrule
Album (DBPedia) & Is the entity discussed in the sentence an example of a album? & 2000 \\ \midrule
Animal (DBPedia) & Is the entity discussed in the sentence an example of a animal? & 2000 \\ \midrule
Artist (DBPedia) & Is the entity discussed in the sentence an example of a artist? & 2000 \\ \midrule
Athlete (DBPedia) & Is the entity discussed in the sentence an example of a athlete? & 2000 \\ \midrule
Building (DBPedia) & Is the entity discussed in the sentence an example of a building? & 2000 \\ \midrule
Company (DBPedia) & Is the entity discussed in the sentence an example of a company? & 2000 \\ \midrule
Educational institution (DBPedia) & Does the sentence discuss a school, university, or college? & 2000 \\ \midrule
Film (DBPedia) & Is the entity discussed in the sentence an example of a film? & 2000 \\ \midrule
Mean of transportation (DBPedia) & Does the sentence discuss a car, ship, train, or plane? & 2000 \\ \midrule
Natural place (DBPedia) & Is the entity discussed in the sentence an example of a natural landscape or environment? & 2000 \\ \midrule
Office holder (DBPedia) & Is the entity discussed in the sentence an example of a office holder? & 2000 \\ \midrule
Plant (DBPedia) & Is the entity discussed in the sentence an example of a plant? & 2000 \\ \midrule
Village (DBPedia) & Is the entity discussed in the sentence an example of a village? & 2000 \\ \midrule
Written work (DBPedia) & Is the entity discussed in the sentence a writing? & 2000 \\ \midrule
Architect (FewRel) & Does the text mention an architect? & 600 \\ \midrule
Composer (FewRel) & Does the text mention a musical composer? & 600 \\ \midrule
Country (FewRel) & Does the text mention a country? & 600 \\ \midrule
Developer (FewRel) & Does the text mention development? & 600 \\ \midrule
Director (FewRel) & Does the text mention a film director? & 600 \\ \midrule
Distributor (FewRel) & Does the text mention a film distributor? & 600 \\ \midrule
Father (FewRel) & Does the text mention a father? & 600 \\ \midrule
Genre (FewRel) & Does the text mention the genre of a song or artist? & 600 \\ \midrule
Instrument (FewRel) & Does the text mention an instrument? & 600 \\ \midrule
League (FewRel) & Does the text mention a sports competition, league, or division? & 600 \\ \midrule
Military Branch (FewRel) & Does the text mention a military branch? & 600 \\ \midrule
Movement (FewRel) & Does the text mention an art movement? & 600 \\ \midrule
Occupation (FewRel) & Does the text mention a professional occupation? & 600 \\ \midrule
Participating Team (FewRel) & Does the text mention a sports team? & 600 \\ \midrule
Platform (FewRel) & Does the text mention an online platform? & 600 \\ \midrule
Sibling (FewRel) & Does the text mention a sibling? & 600 \\ \midrule
Successful Candidate (FewRel) & Does the text mention an election winner? & 600 \\ \midrule
Taxonomy Rank (FewRel) & Does the text mention a taxonomy class of animals? & 600 \\ \midrule
Tributary (FewRel) & Does the text mention a tributary? & 600 \\ \midrule
Winner (FewRel) & Does the text mention a competition winner? & 600 \\ \midrule
YouTube Spam Detection & Does the comment ask the user to check out another video? & 1836 \\
\end{xltabular}
}
\newpage
\section{Prompts}\label{appendix:prompts}
\begin{figure}[h!]
    \centering
    \includegraphics[scale=0.45]{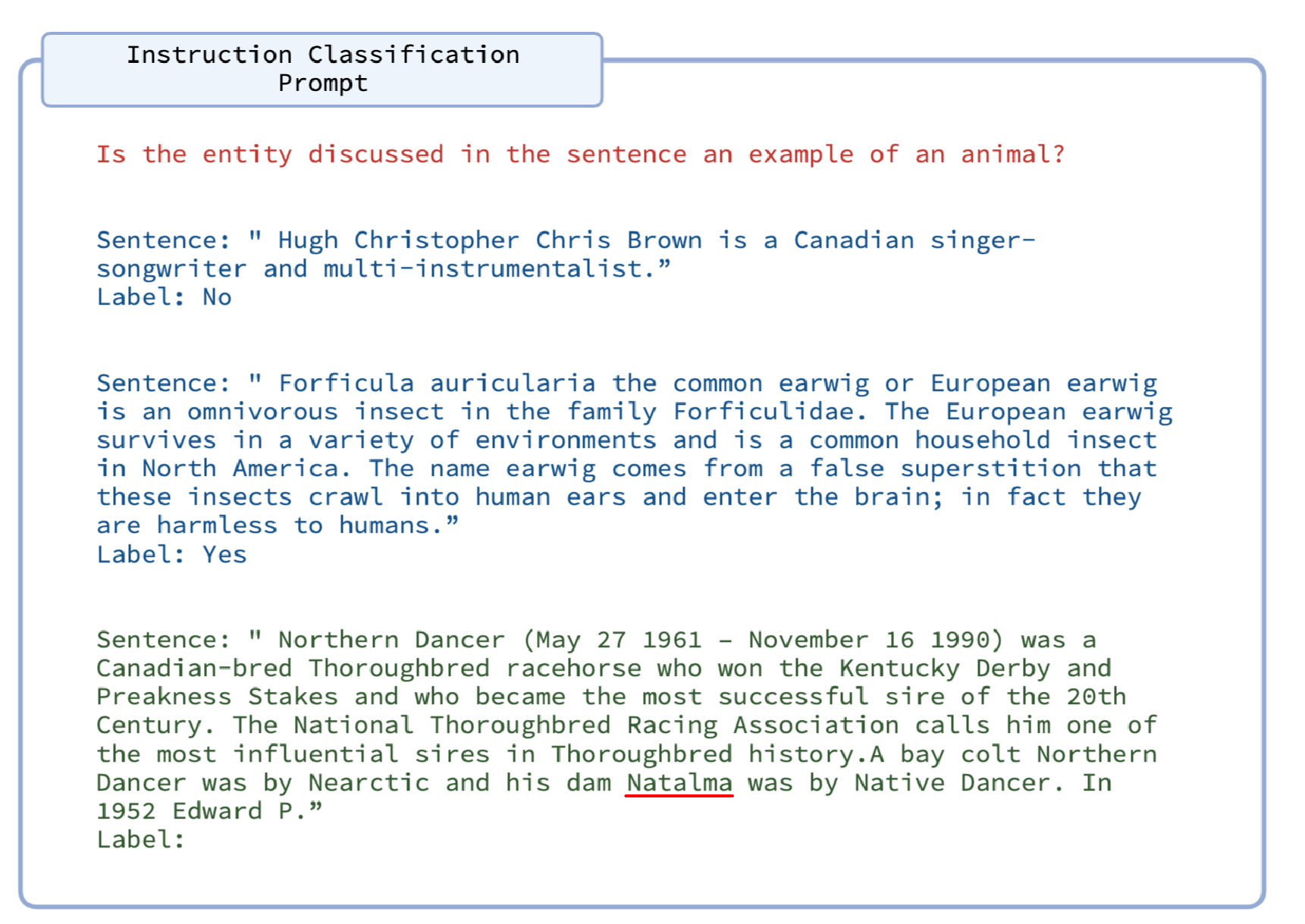}
    \caption{An example of the instruction classification prompt for the dbpedia\_animal task, with two in-context demonstrations. Here, the task instructions are in red, the in-context demonstrations are in blue, and the sample for which we want a label is in green.}
    \label{fig:prompt:instruction_classification}
\end{figure}

We construct a prompt for each task by randomly selecting three examples of each class to use as in-context demonstrations~\cite{perez2021true}. We manually defined ``instructions'' for each task. An example of an abridged prompt is shown in Figure \ref{fig:prompt:instruction_classification}. 
\newpage\section{Synthetics}\label{appendix:synthetics}
We conduct synthetic experiments which provide additional insights on \name. 

For our setup, we create two equal clusters of data of $500$ points each, $C_1$ and $C_2$, in $\mathbb{R}^2$. We assign labels to the points in each cluster i.i.d. according to a probability $p$, where $\Pr(y = 1 | x \in C_1) = p$ and $\Pr(y = 1 | x \in C_2) = 1 - p$. When $p= 0.5$, both the clusters have a uniform label distribution (non-smooth) while $p=1$ ensures each cluster has one class (smooth). We set $k = 20$, $\tau_+ = P(\lambda_i = 1)$ and $\tau_- = P(\lambda_i = -1)$.

\begin{figure*}[th!]
    \centering
    \begin{subfigure}[b]{0.32\textwidth}
         \centering
    \includegraphics[width=\textwidth]{{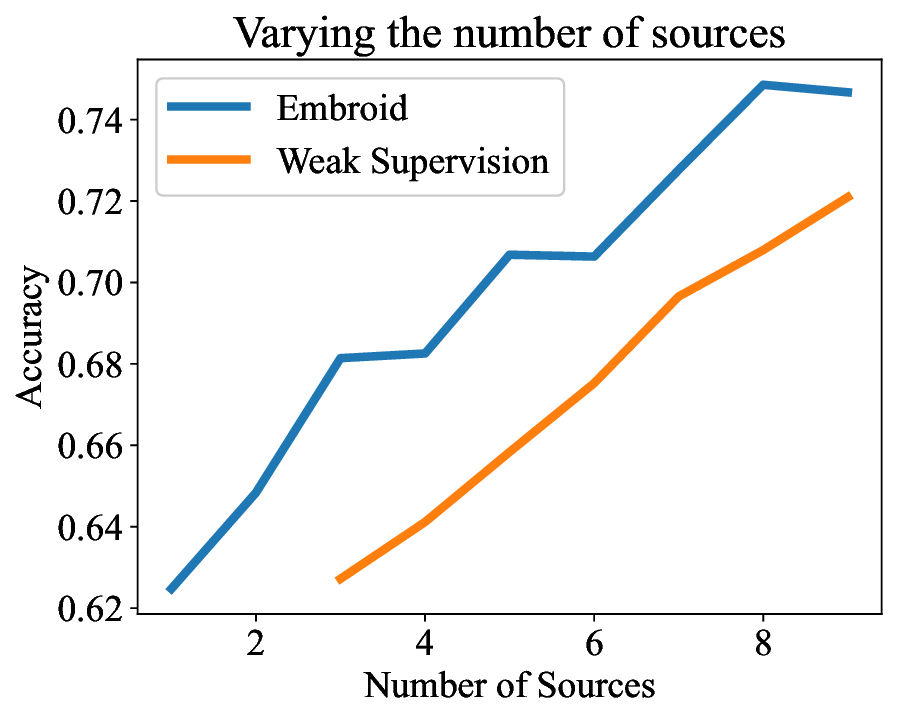}}
    \caption{}
    \label{fig:synthetic_lift_over_ws}
     \end{subfigure}
     \hfill 
    \begin{subfigure}[b]{0.32\textwidth}
         \centering
    \includegraphics[width=\textwidth]{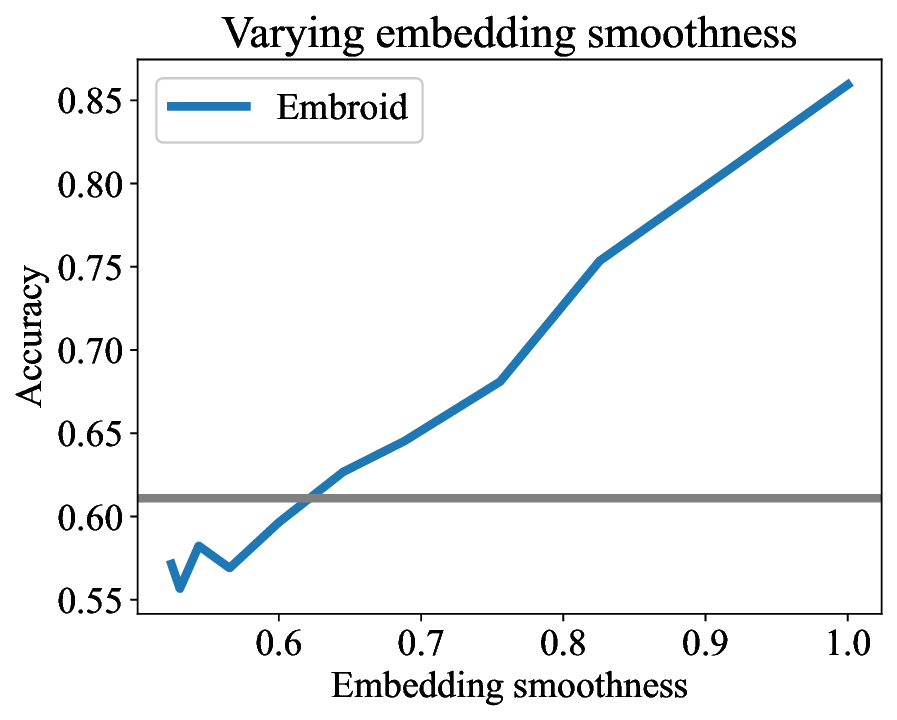}
    \caption{}
    \label{fig:synthetic_embedding_smoothness}
     \end{subfigure}
     \hfill
     \begin{subfigure}[b]{0.32\textwidth}
         \centering
    \includegraphics[width=\textwidth]{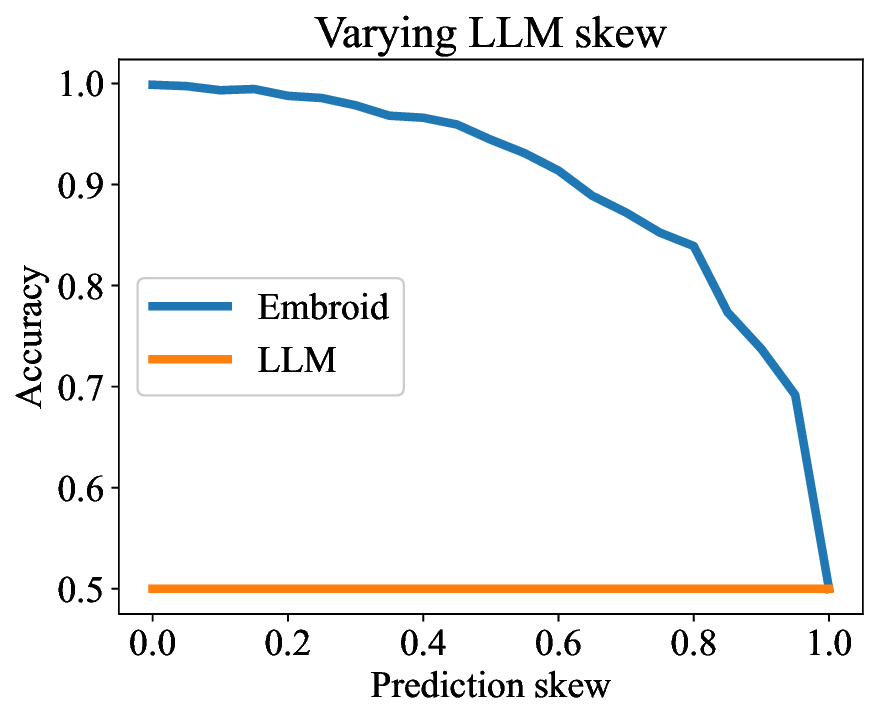}
    \caption{}
    \label{fig:synthetic_skew}
     \end{subfigure}
         \caption{Synthetic experiments. (a) Comparison of \name to weak supervision by varying number of LLM sources. Increasing sources consistently improves both \name and WS and the gains remain constant. (b) \name's performance as embedding smoothness (eq.~\eqref{eq:smoothness}) varies. \name's accuracy linearly improves as a function of embedding smoothness. (c) \name's performance with varying probability of LLM being incorrect in $C_1$. As the LM becomes incorrect, the cluster becomes less homogeneous and this degrades \name's performance.}
\end{figure*}

\paragraph{Improvement over weak supervision} We show that \name offers improvement over methods that only use $\lambdab$. We fix $p = 0.8$ and $\beta_i = \Pr(\lambda_i = y) = 0.6$ for each $i \in [m]$. We compare \name against the standard weak supervision approach from~\cite{fu2020fast}, which requires $m \ge 3$, in Figure~\ref{fig:synthetic_lift_over_ws}.

\paragraph{Smoothness} \name's performance depends on the smoothness of the embedding as defined in eq.~\eqref{eq:smoothness}. We consider one LM prediction $m = 1$ and vary the smoothness $p$ from $0.5$ to $1.0$ and generate predictions using $\beta_i = 0.6$. Figure~\ref{fig:synthetic_embedding_smoothness} exhibits that \name's accuracy is positively correlated with the embedding smoothness.

\paragraph{Base prediction accuracy} Finally, we show that \name's performance depends on the base prediction accuracy, $\beta_i$. We consider $m = 1$ and set $p = 0$, $\beta_i = 0.8$. We use a parameter $\rho$ to denote the probability that $\lambda_i$ is incorrect on points in cluster $C_1$. As $\rho$ varies from $0$ to $1$, the predictions of $\lambda_i$ become biased towards $1$ and effectively reduces $\beta_i$ to $0.5$. 
In Figure~\ref{fig:synthetic_skew}, we observe that as the base prediction accuracy decreases, \name's performance decreases and eventually goes below the base LM performance. 

\section{Ablation}\label{sec:appendix:hyperparameters}
We perform additional ablations of \name. We focus on two aspects: 
\begin{itemize}
    \item The role of $\tau^-$/$\tau^+$.
    \item The impact of weak supervision in combining $\svote{j}$.
\end{itemize}

\subsection{Ablation over $\tau$}
\begin{figure}
    \centering
    \includegraphics[scale=0.5]{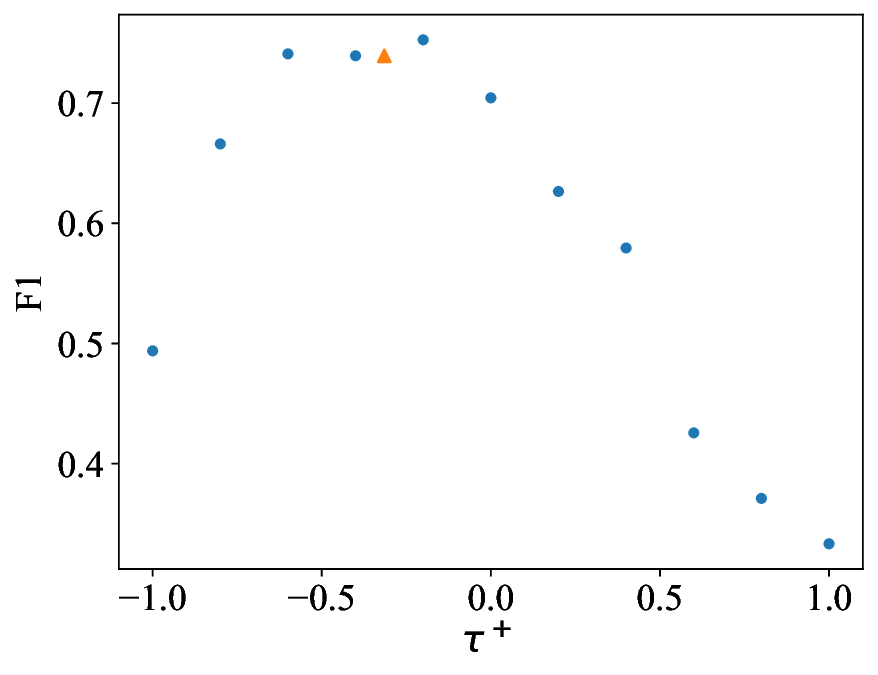}
    \caption{We analyze how F1 changes for different settings of $\tau^+_i$/$\tau^-_i$ for GPT-JT on a task. Observe that setting $\tau^+_i = \tau^-_i = \mathbb{E}[\lambda_i]$ (denoted as the orange triangle) produces close-to-optimal performance.}
    \label{fig:tau_ablation}
\end{figure}

In our experiments, we set $\tau^+_i = \tau^-_i = \mathbb{E}[\lambda_i]$, or the average vote of source $\lambda_i$. This has the following effect on the neighborhood vote $\svote{j}[i]$ for source $\lambda_i$ under $E_j$: 
\begin{itemize}
    \item When the average vote for a source $\lambda_i$ in a neighborhood under $E_j$ for $x$ is more negative than the average overall vote for a source, then $\svote{j}[i](x) = -1$.
    \item When the average vote for a source $\lambda_i$ in a neighborhood under $E_j$ for $x$ is more positive than the average overall vote for a source, then $\svote{j}[i](x) = 1$.
\end{itemize}
In general, we find that this setting provides good performance, while requiring no additional tuning or validation. For example, the Figure \ref{fig:tau_ablation} below compares a setting of $\tau^+_i$/$\tau^-_i$ against the F1 score for GPT-JT on ag\_news\_business. 

\subsection{Role of weak supervision aggregation}
We quantify the extent to which performance gains are derived from (1) the computation of $\svotes$, as opposed to (2) the use weak-supervision~\cite{fu2020fast} to combine $\svotes$ and $\lambdab$. Specifically, we replace Equation \ref{eq:main} with a simple majority vote classifier which combines the original prediction  with the computed neighborhood votes. 

\begin{table}[t!]
    \centering
    \begin{tabularx}{\textwidth}{lccc}
    LM & Base prompt &  Majority vote aggregation  & \name \\
    \toprule
     j1-jumbo & 0.498 & 0.569 & 0.604 \\
    openai\_text-davinci-003 & 0.806 & 0.844 & 0.855 \\
    bloom-7b1 & 54.7 & 61.2 & 64.7 \\
    opt-6.7b & 48.2 & 56.1 & 59.8 \\
    GPT-JT-6B-v1 & 67.8 & 73.2 & 75.1 \\
    \bottomrule
    \end{tabularx}
    \caption{We evaluate how \name compares to a majority vote aggregation over neighborhood vote vectors. We report macro-F1 average across all tasks and prompts, mirroring results in Table \ref{tab:embroid_improvement}.}
    \label{tab:embroid_ablate_ws}
\end{table}
\newpage
\section{Experiments}

\subsection{Implementation details}

\paragraph{Compute} Inference for API-access models (e.g., GPT-3.5 and J1-Jumbo) were run using the HELM API~\cite{liang2022holistic}. Inference for open source models (OPT, GPT-JT, and Bloom) were run using the Manifest library~\cite{orr2022manifest} on 40GB A100 NVIDIA GPU machines.

\paragraph{Hyperparameters} \name was run with $k = 10$, $\tau^{+}_i = P(\lambda_i = 1)$, and $\tau^{-}_i = P(\lambda_i = -1)$

\paragraph{API-model tasks} Due to cost constraints, we study the API access models (GPT-3.5 and J1-Jumbo) on a subset of tasks. These are: 
\begin{itemize}
    \item \texttt{ag\_news\_world}
    \item \texttt{ag\_news\_sports}
    \item \texttt{dbpedia\_educational institution}
    \item \texttt{dbpedia\_athletechemprot\_regulator}
    \item \texttt{chemprot\_upregulator}
    \item \texttt{rct\_objective}
    \item \texttt{rct\_methods}
    \item \texttt{CUAD\_Audit Rights}
    \item \texttt{CUAD\_Non-Compete}
    \item \texttt{learned\_hands\_HE}
    \item \texttt{learned\_hands\_HO}
    \item \texttt{few\_rel\_architect}
    \item \texttt{few\_rel\_league} 
\end{itemize}

\subsection{Robustness across models}
\begin{figure}[h!]
    \centering
    \includegraphics[scale=0.5]{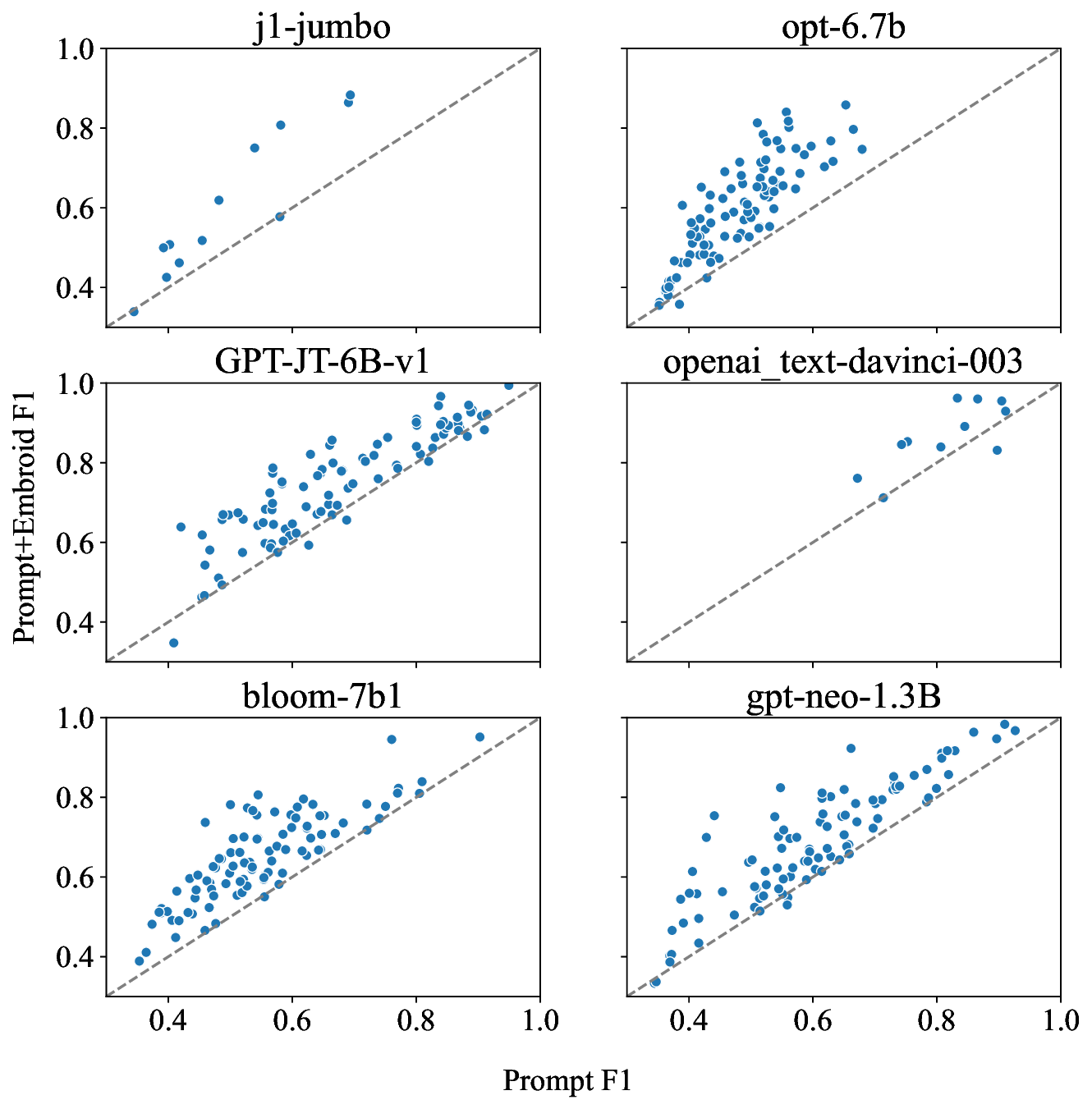}
    \caption{We visualize the improvement from \name over the base prompt for each model's tasks. All models except for gpt-neo-1.3B were run thrice per task, with each run using different in-context samples for the prompt. Each dot corresponds to a task. The x-axis measures the average macro F1 of the base prompt, and the y-axis measures the average macro F1 of \name (across all runs). Because GPT-3.5 and J1-Jumbo are studied on a subset of 12 tasks, there are fewer dots in the plot.}
    \label{fig:all_model_improvements}
\end{figure}

We provide the results for each LM on each task as CSV files in the supplemental attachment. We visualize the improvements in Table \ref{tab:embroid_improvement} below, by plotting the original prompt performance on the x-axis, and the performance after \name on the y-axis (Figure \ref{fig:all_model_improvements}).

\subsection{Comparison to AMA}
\begin{figure}[h!]
    \centering
    \includegraphics[scale=0.5]{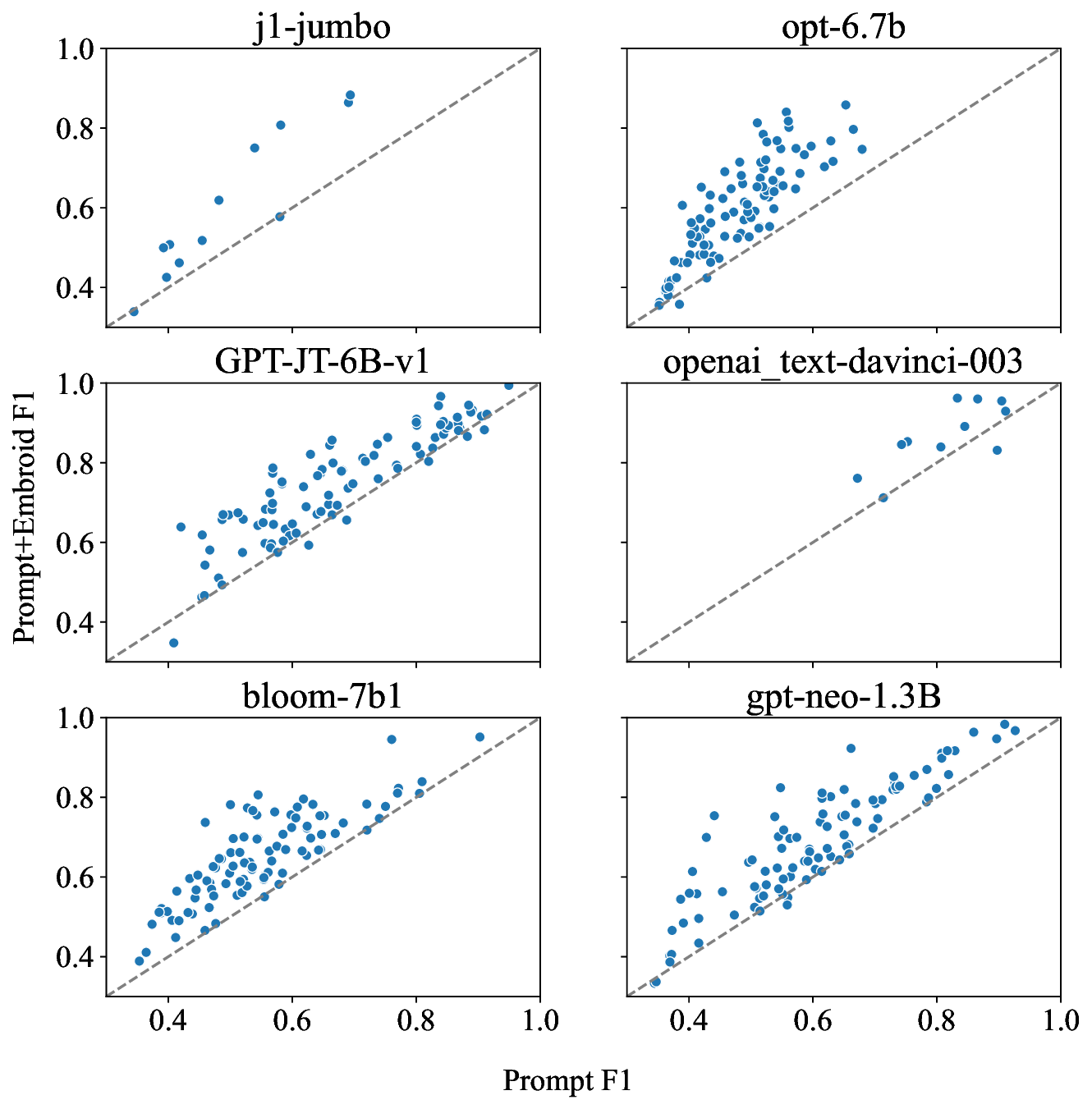}
    \caption{We visualize the improvement from \name over AMA~\cite{arora2022ask} for each model's tasks. Each dot corresponds to a task. The x-axis measures the average macro F1 of AMA, and the y-axis measures the average macro F1 of \name. Because GPT-3.5 and J1-Jumbo are studied on a subset of 12 tasks, there are fewer dots in the plot.}
    \label{fig:embroid_3_v_ama}
\end{figure}

We visualize the improvements in Table \ref{tab:embroid_aggregation} below, by plotting the performance of AMA on the x-axis, and the performance of \name-3 on the y-axis (Figure \ref{fig:embroid_3_v_ama}).

\subsection{Chain-of-thought experiments}
We compare \name to chain-of-thought (CoT) ~\cite{wei2022chain} on the following tasks:
\begin{itemize}
    \item \texttt{ag\_news\_business}
    \item \texttt{ag\_news\_sports}
    \item \texttt{CUAD\_Affiliate License-Licensee}
    \item \texttt{CUAD\_Audit Rights}
    \item \texttt{dbpedia\_album}
    \item \texttt{dbpedia\_building}
    \item \texttt{few\_rel\_architect}
    \item \texttt{few\_rel\_country}
    \item \texttt{learned\_hands\_HO}
    \item \texttt{learned\_hands\_MO} 
\end{itemize}

We manually write logical chains for each prompt. Because CoT primarily succeeds on ``large'' models~\cite{wei2022chain}, we focus our experiments on GPT-3.5.

\end{document}